\documentclass{article}

\usepackage{microtype}
\usepackage{graphicx}
\usepackage{subcaption}
\usepackage{booktabs} 
\usepackage[most]{tcolorbox}
\usepackage{enumitem}
\usepackage{pifont}

\usepackage{hyperref}

\definecolor{darkgreen}{RGB}{0, 130, 0}

\definecolor{darkblue1}{RGB}{0, 130, 200}
\definecolor{darkblue2}{RGB}{0, 100, 160}
\definecolor{darkblue3}{RGB}{0, 65, 130}
\definecolor{darkblue4}{RGB}{0, 25, 75}
\definecolor{darkblue5}{RGB}{0, 10, 40}

\usepackage{xcolor}

\definecolor{darkgreen1}{RGB}{0, 145, 0}

\definecolor{darkgreen2}{RGB}{0, 120, 0}

\definecolor{darkgreen3}{RGB}{0, 95, 0}

\definecolor{darkgreen4}{RGB}{0, 70, 0}

\definecolor{darkgreen5}{RGB}{0, 45, 0}

\usepackage[preprint]{icml2026}

\usepackage{amsmath}
\usepackage{amssymb}
\usepackage{mathtools}
\usepackage{amsthm}
\usepackage{graphicx}
\usepackage{subcaption}
\usepackage{soul}
\usepackage{makecell} 
\usepackage[capitalize,noabbrev]{cleveref}
\usepackage{booktabs}
\usepackage{multirow}
\usepackage{makecell}
\usepackage{graphicx}
\theoremstyle{plain}

\theoremstyle{definition}

\theoremstyle{remark}

\usepackage[textsize=tiny]{todonotes}
\usepackage{colortbl}
\usepackage{diagbox}
\usepackage{xcolor}
\usepackage{caption}
\usepackage{colortbl}
\usepackage{xcolor}
\usepackage{diagbox}
\definecolor{lightgreen1}{rgb}{0.97, 1.00, 0.97}
\definecolor{lightgreen2}{rgb}{0.92, 0.98, 0.92}
\definecolor{lightgreen3}{rgb}{0.84, 0.95, 0.84}
\definecolor{lightgreen4}{rgb}{0.74, 0.91, 0.74}
\definecolor{lightgreen5}{rgb}{0.64, 0.86, 0.64}
\definecolor{lightgreen6}{rgb}{0.54, 0.81, 0.54}

\definecolor{lightorange1}{rgb}{1.00, 0.98, 0.95}
\definecolor{lightorange2}{rgb}{1.00, 0.95, 0.85}
\definecolor{lightorange3}{rgb}{1.00, 0.90, 0.70}
\definecolor{lightorange4}{rgb}{1.00, 0.85, 0.55}
\definecolor{lightorange5}{rgb}{1.00, 0.80, 0.40}
\definecolor{lightorange6}{rgb}{1.00, 0.75, 0.30}

\definecolor{lightpurple1}{rgb}{0.985, 0.97, 1.00}
\definecolor{lightpurple2}{rgb}{0.96, 0.92, 1.00}
\definecolor{lightpurple3}{rgb}{0.93, 0.84, 1.00}
\definecolor{lightpurple4}{rgb}{0.87, 0.74, 1.00}
\definecolor{lightpurple5}{rgb}{0.81, 0.64, 1.00}
\definecolor{lightpurple6}{rgb}{0.75, 0.54, 1.00}

\definecolor{lightred1}{rgb}{1.00, 0.97, 0.97}
\definecolor{lightred2}{rgb}{1.00, 0.92, 0.92}
\definecolor{lightred3}{rgb}{1.00, 0.84, 0.84}
\definecolor{lightred4}{rgb}{1.00, 0.74, 0.74}
\definecolor{lightred5}{rgb}{1.00, 0.64, 0.64}
\definecolor{lightred6}{rgb}{1.00, 0.54, 0.54}
\definecolor{lightcyan1}{rgb}{0.97, 1.00, 1.00}
\definecolor{lightcyan2}{rgb}{0.92, 0.98, 0.98}
\definecolor{lightcyan3}{rgb}{0.84, 0.95, 0.96}
\definecolor{lightcyan4}{rgb}{0.74, 0.91, 0.94}
\definecolor{lightcyan5}{rgb}{0.64, 0.87, 0.92}
\definecolor{lightcyan6}{rgb}{0.54, 0.83, 0.90}
\icmltitlerunning{Submission and Formatting Instructions for ICML 2026}
\definecolor{hiccup}{HTML}{001473}
\usepackage{tcolorbox}
\usepackage{pifont} 
\begin{document}
\setlength{\abovedisplayskip}{3pt plus 1pt minus 1pt}
\setlength{\belowdisplayskip}{3pt plus 1pt minus 1pt}
\setlength{\abovedisplayshortskip}{0pt}
\setlength{\belowdisplayshortskip}{0pt}
\twocolumn[
  \icmltitle{Learning to Explore with Parameter-Space Noise: A Deep Dive into Parameter-Space Noise for Reinforcement Learning with Verifiable Rewards}



  \icmlsetsymbol{equal}{*}

\begin{icmlauthorlist}
    \icmlauthor{Bizhe Bai}{fudan,sii}
    \icmlauthor{Xinyue Wang}{fudan}
    \icmlauthor{Peng Ye}{sail,cuhk}
    \icmlauthor{Tao Chen}{sii,fudan}
\end{icmlauthorlist}

\icmlaffiliation{sii}{Shanghai Innovation Institute, Shanghai, China}
\icmlaffiliation{fudan}{College of Future Information Technology, Fudan University, Shanghai, China}
\icmlaffiliation{sail}{Shanghai AI Laboratory, Shanghai, China}
\icmlaffiliation{cuhk}{The Chinese University of Hong Kong, Hong Kong, China}

\icmlcorrespondingauthor{Bizhe Bai}{bizhe.bai@sii.edu.cn}
\icmlcorrespondingauthor{Tao Chen}{eetchen@fudan.edu.cn}




  \icmlkeywords{Machine Learning, ICML}

  \vskip 0.3in
]



\printAffiliationsAndNotice{}  

\begin{abstract}
Reinforcement Learning with Verifiable Rewards (RLVR) improves LLM reasoning, yet growing evidence indicates an \emph{exploration ceiling}: it often reweights existing solution traces rather than discovering new strategies, limiting gains under large sampling budgets (e.g., \texttt{pass@}256). We address this limitation with \textbf{PSN-RLVR}, which perturbs policy parameters \emph{before} rollout generation to induce temporally consistent, trajectory-level exploration that better preserves long-horizon chain-of-thought coherence than action-space noise. To mitigate the induced sampling–update mismatch, we incorporate \textbf{truncated importance sampling} (TIS), and to avoid expensive KL-based  adaptive noise control , we propose a \textbf{computationally efficient} real-time adaptive noise scheduler driven by a lightweight surrogate that combines semantic diversity with normalized self-certainty. Instantiated on \textbf{GRPO}, a widely used RLVR method, \textbf{PSN-GRPO} consistently expands the effective reasoning capability boundary across multiple mathematical reasoning benchmarks and model families, yielding higher pass@$k$ under large sampling budgets and outperforming prior exploration-oriented RLVR methods (e.g., Pass@$k$-style training) while remaining \emph{orthogonal} and thus composable for additional gains.

\end{abstract}

\section{Introduction}

Reinforcement Learning with Verifiable Rewards (RLVR) has become a central paradigm for improving the reasoning behaviors of Large Language Models (LLMs), delivering substantial gains on domains with automatic correctness signals such as mathematics and code generation ~\cite{wang2024mathvista,ouyang2022training,lambert2025tulu3pushingfrontiers}. In this setting, policy-gradient style algorithms---most notably Proximal Policy Optimization (PPO) ~\cite{schulman2017proximalpolicyoptimizationalgorithms} and its recent variants such as Group Relative Policy Optimization (GRPO) ~\cite{shao2024deepseek}---enable direct optimization against verifiers (e.g., unit tests or symbolic checkers), aligning models with ground-truth correctness and propelling systems like DeepSeek-R1~\cite{shao2024deepseek} to strong empirical performance.
\begin{figure*}[h]
        \centering
        \includegraphics[width=\linewidth]{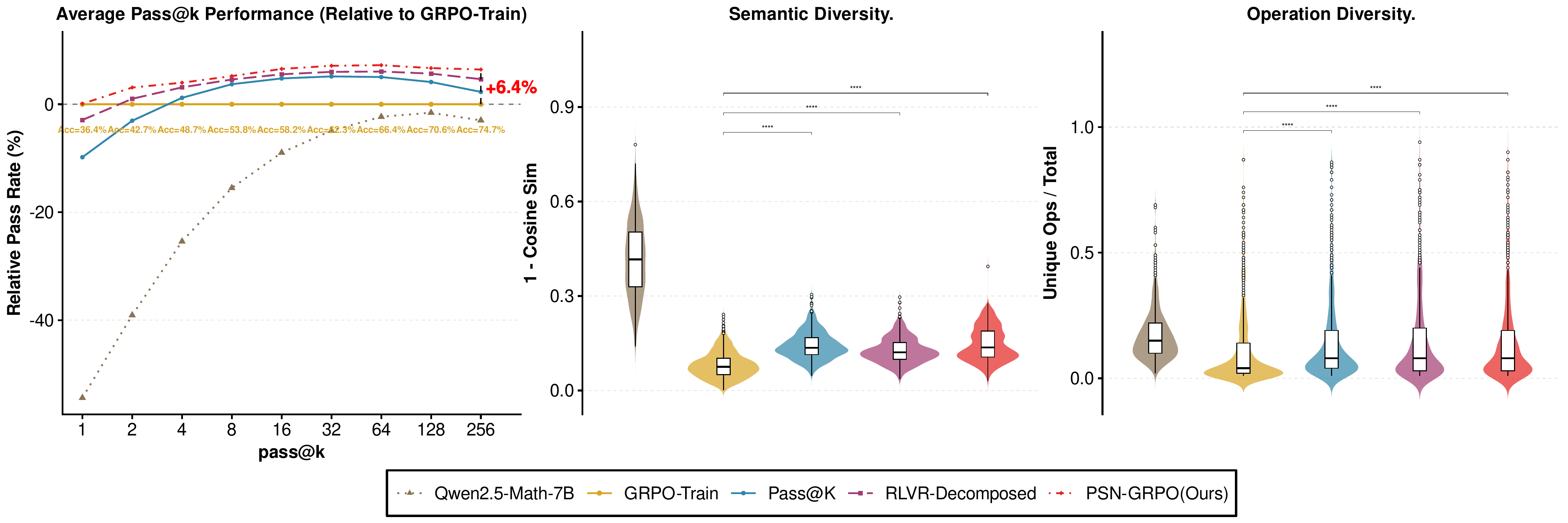}
    \caption{Comparison of reasoning capability boundaries across different RLVR paradigms. (1) Standard RLVR-trained models (GRPO-Train) exhibit a significant reduction in \textbf{semantic diversity} and \textbf{operation diversity} compared to the base model (Qwen2.5-Math-7B). (2) Our proposed method, \textbf{PSN-GRPO}, restores and enhances this diversity, achieving significantly higher semantic and operation variance compared to the GRPO baseline. (3) In terms of reasoning performance, PSN-GRPO is superior to other exploration-focused methods, such as \textbf{Pass@k training}~\cite{Chen2025PasskTF} and \textbf{RLVR-Decomposed}~\cite{zhu2025surprisingeffectivenessnegativereinforcement}, consistently delivering higher \textbf{pass@k} metrics, particularly under large sampling budgets (e.g., $k=128, 256$).}
    \label{fig:different_methods}
\end{figure*}
However, emerging evidence suggests that current RLVR pipelines may face an \emph{exploration ceiling}. Recent analyses~\cite{yue2025does, wu2026invisibleleashrlvrescape} indicate that RLVR tends to improve \emph{sampling efficiency}  (i.e., pass@1) rather than inducing genuinely new reasoning capability boundaries (i.e., pass@256) : the trajectories produced after RLVR are largely contained within (or near) the base model's pretraining distribution. In addition, the RLVR-trained LLM tend to have less semantic diversity and operation diversity compared to the original model~\cite{dang2025assessing} as shown in Figure~\ref{fig:different_methods} (right). Under this view, RLVR behaves primarily as a distributional \emph{reweighting} mechanism---it amplifies pre-existing correct trajectories while rarely discovering qualitatively new solution strategies. This leaves an ``exploration gap'': the inability to reliably traverse regions of the reasoning space that are unlikely under the initial policy but may contain superior or more robust solutions.

Bridging this gap requires balancing exploration~\cite{Sutton1998ReinforcementLA} and exploitation while preserving the long-horizon dependency essential for Chain-of-Thought (CoT) reasoning. Existing approaches generally fall into three categories, each with distinct limitations:
\textbf{1. Action-Space Perturbations (Decoding-Time):} Techniques such as temperature sampling \citep{Renze2024TheEO} or nucleus sampling \citep{Holtzman2019TheCC} inject stochasticity at the token level. However, token-level noise is typically \emph{uncorrelated} across time steps. In multi-step reasoning, small perturbations in each step may accumulate into unstructured noise that degrades global coherence, harming long-horizon reasoning trajectories ~\cite{Renze2024TheEO}, rendering the CoT state-level inconsistent. 
\textbf{2. Objective-Level Regularization:} Methods that modify the training objective—such as entropy bonuses ~\cite{Zhan2025MindYE} or pass@k optimization ~\cite{Chen2025PasskTF}—attempt to force diversity explicitly. \textbf{3. Data Augmentation:} 
While experience augmentation (e.g., self-generated task variants, offline data, or external teacher models)~\cite{liang2025pass1selfplayvariationalproblem,dong2025rlpluscounteringcapabilityboundary,li2025questaexpandingreasoningcapacity} can broaden support but often introduces additional computational cost or reliance on external signals. More discussion is provided in Section~\ref{sec:related_work_explore}.

To overcome these limitations, we propose \textbf{PSN-RLVR}, a parameter-space exploration framework for RLVR that perturbs policy parameters prior to rollout generation, inducing \emph{temporally consistent}, trajectory-level exploration better aligned with long-horizon chain-of-thought reasoning than token-level noise. Beyond introducing PSN to RLVR, we \emph{comprehensively explore} its design space in this setting (Section~4.2), systematically characterizing where to inject noise, how performance scales with noise magnitude, when PSN is preferable to action-space perturbations, and how PSN interacts with existing RLVR exploration techniques. Applying PSN in RLVR poses two unique challenges—off-policy sampling--update mismatch and compute-aware noise control—which we address with two lightweight modules: (i) \textbf{truncated importance sampling (TIS)} to stabilize optimization under rollouts collected from the perturbed sampler, and (ii) a \textbf{real-time adaptive noise scheduler} that replaces expensive KL-based control with a low-overhead surrogate combining semantic diversity and normalized self-certainty. Instantiated on GRPO, PSN-RLVR consistently expands the effective reasoning capability boundary, improving pass@k under large sampling budgets while remaining orthogonal and composable with prior RLVR enhancements.

To the best of our knowledge, this paper presents the \textbf{first systematic study of parameter-space noise} for Large Language Models trained with Verifiable Rewards (RLVR). Our main contributions are:
\begin{itemize}
    \item \textbf{PSN-RLVR: parameter-space exploration for RLVR.} We introduce a parameter-space noise framework for RLVR that perturbs policy parameters before rollout generation to induce temporally consistent, trajectory-level exploration, and instantiate it on GRPO to form \textbf{PSN-GRPO}.
\vspace{-0.5em}
    \item \textbf{Two modules for RLVR-specific challenges.} To address the sampling--update mismatch induced by parameter perturbations, we incorporate \textbf{truncated importance sampling (TIS)} for stable off-policy learning; to avoid expensive KL-based noise control, we propose a \textbf{lightweight real-time adaptive noise scheduler} driven by a surrogate that combines semantic diversity and model's self-certainty.
\vspace{-0.5em}
    
    \item \textbf{Comprehensive exploration of the PSN design space in RLVR.} We conduct extensive experiments and targeted ablations (Section~4.2) that systematically answer key design questions—noise injection location, noise magnitude scaling, robustness across model families, comparisons to action-space noise, and complementarity with other exploration-oriented RLVR methods—demonstrating consistent improvements in high-budget pass@k and reasoning diversity.
\end{itemize}


\section{Related Work}

\subsection{Reinforcement Learning for Reasoning with Verifiable Rewards}
The integration of Reinforcement Learning (RL) into the post-training pipeline of Large Language Models (LLMs) has become a standard paradigm for enhancing performance in domains with objective ground-truth, such as mathematics and coding ~\cite{ouyang2022training, wang2024mathvista, shao2024deepseek}.  More discussion in Appendix~\ref{sec:app_related_work}.


\subsection{The Exploration-Exploitation Boundary in LLMs}
\label{sec:related_work_explore}
Existing approaches generally fall into three categories(1) Action-Space Perturbations (Decoding-Time)  (2) Objective-Level Regularization (3) Data Augmentation. We provide detailed summary of three classes and state their  distinct limitations in following.
First, a large class of methods modulates exploration at \emph{decoding time} by perturbing the action space---e.g., temperature sampling~\cite{Renze2024TheEO}, nucleus/top-$k$ sampling~\cite{Holtzman2019TheCC}, or prompt perturbations \cite{ShurOfry2024GrowingAT}. Although these heuristics can increase diversity, they typically demand careful hyperparameter tuning~\cite{du2025optimizingtemperaturelanguagemodels} and can be sensitive across domains and model families~\cite{Shi2024ATE,qiang2024promptperturbationconsistencylearning,chen2024failuresselfconsistencymultistepreasoning}. More fundamentally, token-level stochasticity is often \emph{locally uncorrelated} across time~\cite{Renze2024TheEO}, so small perturbations can accumulate into unstructured jitter that reduces long-horizon CoT coherence, derailing global logical consistency in difficult reasoning trajectories.

Second, training-time diversification can be pursued by modifying the RLVR objective, such as substituting the pass@1 objective with pass@k objectives~\cite{Chen2025PasskTF,peng2025simkosimplepasskpolicy} in GRPO~\cite{shao2024deepseek}, using output entropy as a proxy for exploration control \cite{cheng2025reasoningexplorationentropyperspective,cui2025entropymechanismreinforcementlearning}, or incorporating negative samples to promote exploration \cite{zhu2025surprisingeffectivenessnegativereinforcement}. While promising, objective-level regularization often depends on proxy signals whose effectiveness can be sensitive to task difficulty and reward sparsity.

Third, another line of work relies on data and experience augmentation to broaden exploration, such as creating new task variations from the model itself~\cite{liang2025pass1selfplayvariationalproblem}, leveraging off-line data~\cite{dong2025rlpluscounteringcapabilityboundary,li2025questaexpandingreasoningcapacity}, or using additional LLM guidance~\cite{jiang2025selectiveexpertguidanceeffective}. Although these approaches can expand the effective support of training, they typically bring additional computation cost, require extra data curation, or introduce external signals beyond the base RLVR loop.

\subsection{Parameter-Space Noise}
\textit{Parameter-Space Noise} (PSN) offers another compelling alternative by perturbing the policy weights rather than the actions. Works such as  \cite{plappert2018parameter} and \cite{fortunato2018noisy} demonstrated in continuous control domains that PSN induces state-dependent exploration: a perturbed policy acts as a distinct "agent" that executes a consistent strategy over an entire episode. While PSN is well-established in robotics~\cite{fortunato2018noisy,Gupta2018MetaReinforcementLO,Hollenstein2022ActionNI,gravell2021robustlearningbasedcontrolbootstrapped}, its application to the discrete, high-dimensional reasoning space of RLVR remains largely unexplored. On the other side, subsequent theoretical work situates such noise-injected policies within the broader paradigm of posterior sampling / randomized value functions, which can induce deep exploration and admits regret guarantees ~\cite{osband2019deep,russo2019worstcase}. 
In particular, variational Thompson-sampling perspectives interpret practical parameter-noise methods  as tractable approximations to posterior sampling over value functions, providing a theoretical foundation for their empirical effectiveness~\cite{aravindan2021stateaware}. \\
A  concurrent work is QERL~\citep{huang2025qerlefficiencyquantizationenhanced}, which focuses on improving RLVR training efficiency via quantization. While they report that the noise introduced by quantization \emph{surprisingly} improved exploration, this phenomenon was observed as  side effect of their efficiency optimization. Consequently, they did not systematically analyze the noise dynamics, optimal injection strategies, or scaling laws required to leverage this effect for reasoning.


\section{Methodology}

\subsection{Preliminaries: Group Relative Policy Optimization}
\label{sec:preliminaries}
We build our framework upon Group Relative Policy Optimization (GRPO)~\citep{shao2024deepseek}, a widely used RLVR method, which optimizes a policy $\pi_\theta$ without a value function. For each query $q \sim P(Q)$, GRPO samples a group of outputs $\{o_i\}_{i=1}^G$ from the old policy $\pi_{\theta_{\text{old}}}$ and optimizes the following objective:
\begin{equation}
\mathcal{J}_{\mathrm{PPO}}(\theta)= \mathbb{E}_{q \sim P(Q),\, o \sim \pi_{\theta_{\text{old}}}(O \mid q)}
\left[\frac{1}{|o|}\sum_{t=1}^{|o|}\ell_t^{\mathrm{clip}}(\theta)\right],
\label{eq:grpo}
\end{equation}
where $\ell_t^{\mathrm{clip}}(\theta)
= \min\!\left(r_t(\theta)A_t,\;
\operatorname{clip}\!\left(r_t(\theta),1-\varepsilon,1+\varepsilon\right)A_t\right)$, and   $r_t(\theta)=\frac{\pi_\theta(o_t \mid q,o_{<t})}{\pi_{\theta_{\text{old}}}(o_t \mid q,o_{<t})}$
where $A_t$ is the advantage computed from relative rewards within the group.
\subsection{\textbf{P}arameter-\textbf{S}pace \textbf{N}oise GRPO (PSN-GRPO)}
\begin{figure}[t]
    \centering
        \includegraphics[width=\linewidth]{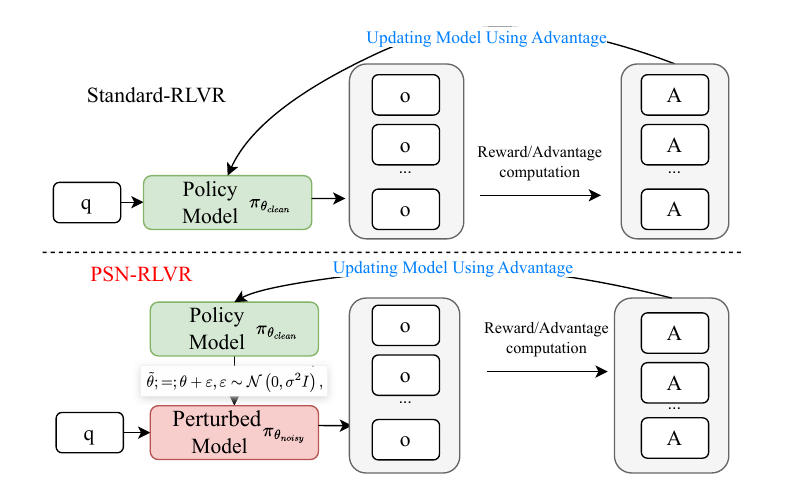}
        \label{fig:framework}
        \caption{Overview of the PSN-RLVR framework compared to Standard-RLVR. The noise-perturbed model $\pi_{\tilde{\theta}}$ generates rollouts to induce temporally consistent, trajectory-level exploration. The resulting reward signals are used to update the clean policy $\pi_{\theta}$}
    \label{fig:main_frame}
\end{figure}

We initialize PSN-RLVR with PSN-GRPO,  an exploration-enhanced  training framework that injects parameter-space noise into the rollout policy to induce temporally consistent exploration while updating the clean policy parameters via policy gradient learning. Because rollouts are generated by a noisy sampler policy, we correct the resulting off-policy mismatch using Truncated Importance Sampling (\textbf{TIS}). Finally, we propose a computationally efficient, real-time scheduler for  noise. The main framework is illustrated in Figure~\ref{fig:main_frame}.

\subsection{Parameter-space noise exploration policy}
\label{sec:param_noise}
Following~\citet{plappert2018parameter}, we explore by perturbing parameters rather than actions/tokens.
At the beginning of each iteration, we apply additive Gaussian noise to the parameter vector of the current policy :
\begin{equation}
\tilde{\theta} ;=; \theta + \varepsilon,
\varepsilon \sim \mathcal{N}\left(0,\sigma^2 I\right),
\end{equation}
which induces a \emph{noisy exploration policy} $\pi_{\tilde{\theta}}$ .
Crucially, $\tilde{\theta}$ is held fixed for the entire rollout, yielding temporally consistent exploration: conditioned on the same prefix $(q,o_{<i})$.

\subsection{Off-Policy Correction via Truncated Importance Sampling}
\label{subsec:imp_sampling}
A distribution mismatch arises because data is collected by $\pi_{\tilde{\theta}}$ but used to train $\pi_{\theta}$. Ignoring this discrepancy yields biased gradient estimates. We correct this via Truncated Importance Sampling (TIS)~\citep{ionides2008truncated}.
We modify the standard GRPO objective by incorporating the importance ratio $w_t$ into the gradient update. The corrected objective is:
\begin{equation}
\mathcal{J}_{\text{PSN}}(\theta) = \mathbb{E}_{q \sim P(Q), o \sim \color{red}{\pi_{\tilde{\theta}}}} \left[ \frac{1}{|o|} \sum_{t=1}^{|o|} w_t \cdot \ell_t^{\text{clip}}(\theta) \right].
\label{eq:psngrpo}
\end{equation}
 To prevent unbounded variance when $\pi_{\tilde{\theta}}$ diverges significantly from $\pi_{\theta}$, we truncate the importance ratio:
\begin{equation}
w_t := \min\!\left(
\frac{\pi_{\theta}\!\left(a_t\right)}
{\pi_{\color{red}{\tilde{\theta}}}\!\left(a_t\right)},
C\right)
\label{eq:tis_c}
\end{equation}
where $C$ is a clipping hyperparameter. This formulation allows the learner to leverage exploratory data from $\pi_{\tilde{\theta}}$ while maintaining training stability.
\subsection{Adaptive Noise Scheduling}
\label{subsec:adaptive_noise}
We would like the exploration policy to remain sufficiently close to the learner to keep off-policy correction stable, while still being different enough to encourage exploration.
We therefore measure how far the noisy policy deviates from the clean policy on the \emph{current} batch and adjust the noise for the \textbf{next} batch. We propose two adaptive scaling variants to dynamically adjust $\sigma$.

\subsubsection{Variant I: Non-Real Scheduler}
Following~\cite{plappert2018parameter}, we adjust $\sigma$ to maintain a target KL divergence, $\delta_{\text{KL}}$, between the clean and noisy policies. After each batch, we compute:
\begin{equation}
d(\pi_{\theta}, \pi_{\tilde{\theta}}) = \mathbb{E}_{s} \left[ \text{KL}\left( \pi_{\tilde{\theta}}(\cdot|s) \parallel \pi_{\theta}(\cdot|s) \right) \right].
\end{equation}
The noise scale for the next iteration, $\sigma_{k+1}$, is updated via:
\begin{equation}
\sigma_{k+1} = \begin{cases} 
\beta \sigma_k & \text{if } d(\pi_{\theta}, \pi_{\tilde{\theta}}) \leq \delta_{\text{KL}} \\ 
\frac{1}{\beta} \sigma_k & \text{otherwise},
\end{cases}
\label{eq:non_real_time_adatpive}
\end{equation}
where $\beta > 1$ is a step constant. However, this retrospective adaptation introduces feedback latency. Given the high variance in problem difficulty within RLVR datasets, this lag can lead to suboptimal noise scheduling: high noise levels necessitated by a difficult problem batch may be inappropriately applied to a subsequent simple problem batch, causing a mismatch that hinders efficient training.
\subsubsection{Variant II: Real-Time Computationally Efficient Scheduler}
To address the feedback lag and computational cost of full rollout sampling, we propose a real-time scheduler based on \textit{semantic diversity} and \textit{model self-certainty}. For each query $q$, we pre-generate two probe rollouts $o^{(1)}, o^{(2)}$ using the clean policy $\pi_{\theta}$ to gauge the model's current exploration needs.
\label{sec:real_time_noise}
\paragraph{Motivation.}
Ideally, accurate noise calibration for the current batch would require computing $d(\pi_{\text{clean}}, \pi_{\text{noisy}})$ \textit{a priori}. However, this is computationally prohibitive. Since rollout generation accounts for 70--80\% of total training time~\cite{cheng2025fastllmposttrainingdecoupled}, a naive approach—sampling a set of rollouts (e.g., $N=8$) to measure divergence, adjusting noise, and then resampling for backpropagation—would nearly double the computational overhead. To circumvent this, we propose a computationally efficient schedule that calibrates noise based on immediate signals of \textit{semantic similarity} and \textit{model self-certainty}. To this end, for each query $q$, we pre-generate two rollouts, $o^{(1)}$ and $o^{(2)}$, using $\pi_{\text{clean}}$ to compute indicators based on semantic embeddings~\cite{dang2025assessing} and self-certainty~\cite{kang2025scalablebestofnselectionlarge}. We explicitly avoid using $KL(\pi_{\text{clean}} || \pi_{\text{noisy}})$ for this adjustment because ``KL between language models notoriously suffers from high variance''~\cite{amini2025betterestimationkullbackleiblerdivergence,fang2025wrongperplexitylongcontextlanguage}, particularly when estimated from only two samples.
\paragraph{Semantic similarity.}
Let $f(\cdot)$ be a fixed sentence embedding model; we compute
\begin{equation}
d_{\mathrm{sem},i}
\;=\cos\!\big(f(o^{(1)}), f(o^{(2)})\big), \overline{d}_{\mathrm{sem},t}=\frac{\sum_{1}^{B}d_{\mathrm{sem},i}}{B}
\label{eq:emb_sim}
\end{equation}
where B is batch size, and $\overline{d}_{\mathrm{sem},t}$ is mean of semantic similarity of current batch. Higher similarity indicates models  tend to generate similar rollouts, signaling a need for greater exploration (i.e., larger parameter noise).
\paragraph{Self-Certainty.}
We quantify \emph{distributional self-certainty} by measuring how far the model's token-level predictive distribution departs from a uniform prior over the vocabulary. \ul{The key intuition is that a sharper (more concentrated) distribution corresponds to greater confidence.} Concretely, for a query $q$ and a generated completion $o=(o_1,\ldots,o_{|o|})$, let $U$ be the uniform distribution on the vocabulary $V$. We define self-certainty as the mean, across decoding steps, of the KL divergence from $U$ to the model distribution $p_{\pi_\theta}$:
\begin{equation}
\mathrm{Self\text{-}certainty}(o \mid q)
= \frac{1}{|o|}\sum_{i=1}^{|o|}
\mathrm{KL}\!\left(U \,\big\|\, p_{\pi_\theta}(\cdot \mid q, o_{<i})\right),
\end{equation}
where $o_{<i}$ are the previously generated tokens and $p\left(j \mid q, o_{<i}\right)$ is the model's predicted probability for token $j$ at step $i$. Higher self-certainty values indicate greater confidence. Larger values imply stronger concentration of probability mass, i.e., greater deviation from uniformity and thus requiring more exploration.
We normalize the batch-averaged self-certainty $\text{SC}_t$ to $[0,1]$ using a global history buffer of running extrema ($S_{\min}, S_{\max}$):
\begin{equation}
\overline{\text{SC}}^{\text{norm}}_t = \text{clip}\left( \frac{\text{SC}_t - S_{\min}}{S_{\max} - S_{\min} + \epsilon}, 0, 1 \right).
\end{equation}
\paragraph{Update Rule.} We define a composite indicator $\text{Ind}_t = \overline{\text{SC}}^{\text{norm}}_t + \overline{d}_{\text{sem}}$ for current batch. A high $\text{Ind}_t$ implies the model is both confident and producing semantically similar outputs, signaling a need for stronger perturbations. We compare $\text{Ind}_t$ to its historical mean $\overline{\text{Ind}}$ to update $\sigma$:
\begin{equation}
\sigma_{k} = \begin{cases} 
\beta \sigma_{k-1} & \text{if }  \overline{\text{Ind}} \leq \text{Ind}_t  \quad \\ 
\frac{1}{\beta} \sigma_{k-1} & \text{otherwise} \quad 
\end{cases}
\label{eq:real_time_update}
\end{equation}
\paragraph{Compute overhead.}
Variant~II uses two probe generations per query. Empirically,  we observe a smaller end-to-end throughput reduction of \(\approx 8\%\) relative to fixed-\(\sigma\) PSN under identical hardware. We attribute the gap to \emph{generation-time imbalance}~\cite{he2025historyrhymesacceleratingllm}. More discussion is illustrated in Appendix~\ref{sec:computation_overhead}.

\section{Experiments}
\subsection{Experimental setup}
\paragraph{Training, Models, and Datasets.}
We adopt GRPO~\cite{shao2024deepseek} as the default RLVR training algorithm. Unless otherwise specified, all experiments utilize the standard configuration of SimpleRL-Zoo\footnote{https://github.com/hkust-nlp/simpleRL-reason}~\cite{zeng2025simplerlzooinvestigatingtamingzero}, including learning rate, batch size, rollout count, and maximum sequence length; detailed hyperparameters are provided in Appendix~\ref{app:train_detail}. We test our method on Qwen2.5-Math-7B~\cite{yang2024qwen25mathtechnicalreportmathematical} and Qwen2.5-7B~\cite{qwen2025qwen25technicalreport}. Following the SimpleRL-Reason protocol~\cite{zeng2025simplerlzooinvestigatingtamingzero}, training data is sampled from the NuminaMath dataset~\cite{numina_math_datasets}, which is derived from GSM8K~\cite{gsm8k} and MATH~\cite{hendrycksmath2021}. Unless explicitly stated otherwise, all reported results rely on the Qwen2.5-Math-7B model trained under these default settings.
\paragraph{Evaluation Protocol.}
Evaluations are conducted on a variety of reasoning benchmarks: AIME 2024, AIME 2025, AMC 2023~\cite{aime24}, OlympiadBench~\cite{he2024olympiadbench}, and Minerva Math~\cite{lewkowycz2022solving}. To ensure a fair and consistent comparison, all our evaluation framework is built upon the open-source \texttt{simpleRL-reason} codebase \cite{zeng2025simplerlzooinvestigatingtamingzero}. To obtain a comprehensive evaluation, we adopt the unbiased pass@$K$ metric with $K$ up to 256, computed as $\text{pass}@K := \mathbb{E}_{x \sim \mathcal{D}} [1 - \binom{n-c}{K} / \binom{n}{K}]$, where $c$ denotes the number of correct completions out of $n$ generated responses following~\cite{peng2025simkosimplepasskpolicy}. To reduce evaluation variance, we set $n = 300$ for all datasets. Unless otherwise specified, we use a decoding temperature $T = 0.9$ for all evaluation tasks.
\paragraph{Metrics representing reasoning capability boundary.}
We use two metrics for representing reasoning capability boundary of RLVR (1) pass@k  following by~\cite{yue2025does,peng2025simkosimplepasskpolicy,Chen2025PasskTF} (2) semantic embedding diversity and operation  diversity between LLM's rollouts following by~\cite{dang2025assessing}. Semantic Diversity: the average cosine similarity between the text embeddings of the rollouts, computed using  Bert-MLM\_arXiv-MP-class\_zbMath~\cite{steinfeldt2024bert_mlm_arxiv_mp_class_zbmath_model}. Operation Diversity: group rollouts by the sequence of arithmetic operations performed and measure the fraction of unique operation sequences.



\subsection{Results and Q\&As}
\label{subsec:results_qa}

We present our results via the following Q\&As.

\ul{\textbf{Q1: Does PSN-RLVR expand the reasoning capability boundary?}}
\begin{tcolorbox}[
    colback=blue!5!white,  
    colframe=blue!5!white, 
    boxsep=1pt,            
    left=2pt,              
    right=2pt,             
    top=2pt,               
    bottom=2pt,            
    arc=2pt                
]
A: Yes, PSN-RLVR achieves higher pass@k under large sampling budgets and increased rollouts semantic diversity and operation diversity.
\end{tcolorbox}

\begin{figure}[]
    \centering
    \includegraphics[width=\linewidth]{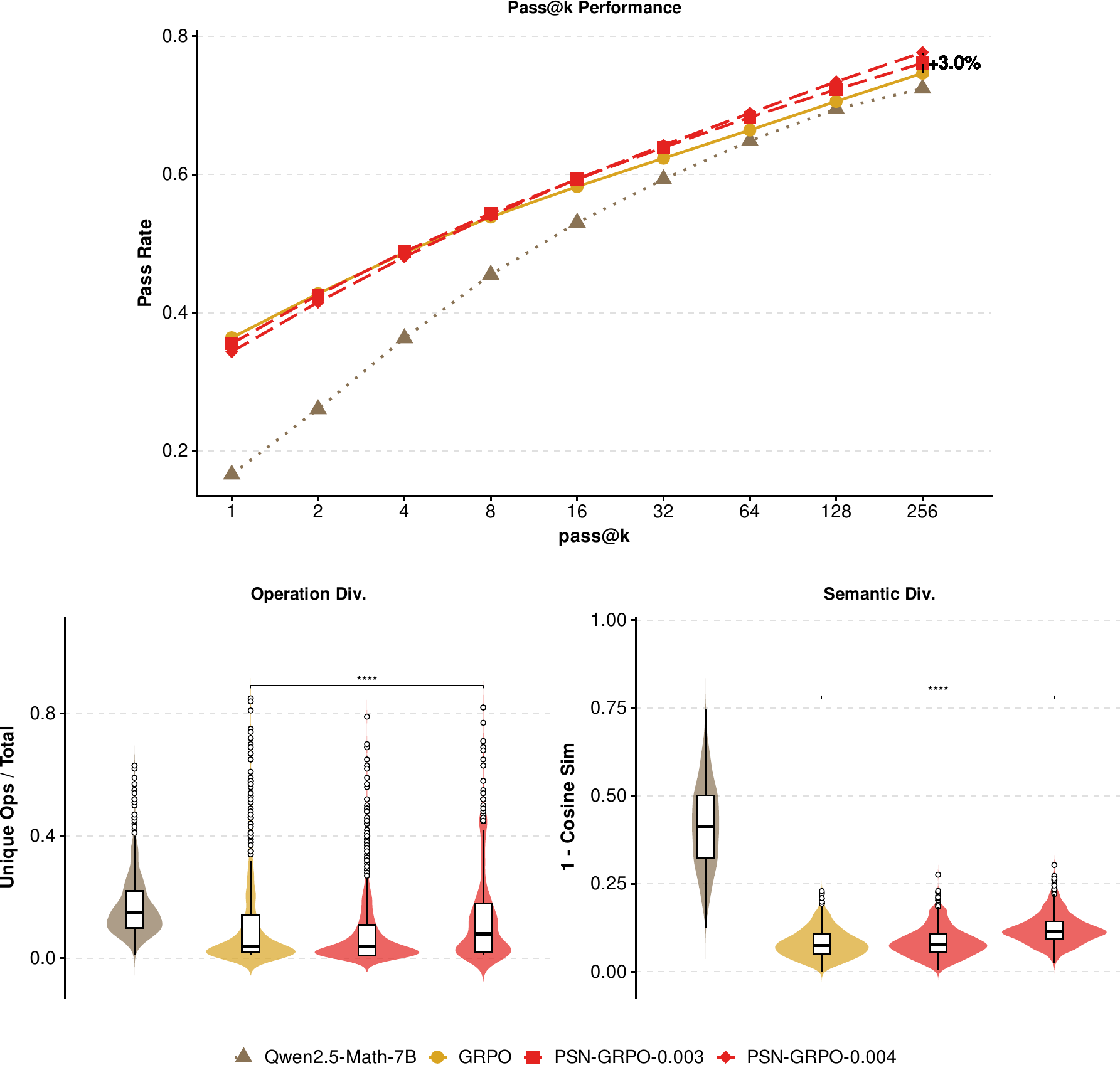}
\caption{ We compare the pass@$k$ performance (top), semantic diversity , and operation diversity  of PSN-GRPO against the standard GRPO baseline on Qwen2.5-Math-7B. PSN-GRPO achieves superior performance at large sampling budgets ($k \ge 16$). This gain is strongly correlated with increased semantic and operational diversity in generated trajectories~\cite{dang2025assessing}.}    \label{fig:q1}
\end{figure}
We initialize the experiment with  PSN-GRPO with \textbf{Qwen2.5-Math-7B} model. Figure~\ref{fig:q1} summarize the average performance across the evaluated datasets. We find that naive PSN-GRPO(without TIS and adaptive noise)  outperforms the standard GRPO baseline when $k$ is large (e.g., from $k=16$ to $k=256$). While standard RLVR improves selection efficiency among pre-existing trajectories, PSN effectively expands the reasoning search space. As shown in Figure~\ref{fig:q1}, this performance boost correlates with significantly higher semantic and operation diversity compared to the baseline, confirming that PSN induces genuinely new reasoning modes rather than simply reweighting the pre-training distribution. Meanwhile, the naive PSN do hurt the pass rate when k is small. This will be mitigated through adaptive nosie mechanism and answered in Question 6.

\ul{\textbf{Q2: Does PSN-RLVR generalize to other models, and where Should Noise Be Injected?}}
\label{q:where_to_injectj}
\begin{tcolorbox}[
    colback=blue!5!white,  
    colframe=blue!5!white, 
    boxsep=1pt,            
    left=2pt,              
    right=2pt,             
    top=2pt,               
    bottom=2pt,            
    arc=2pt                
]
A: Yes. We demonstrate that PSN is model-agnostic by training  on Qwen2.5-7B and Qwen3-4B. Furthermore, both theoretical intuition and empirical results identify MLP layers as the optimal injection site for maximizing the reasoning capability boundary.
\end{tcolorbox}
\begin{figure}[]
    \centering
    \includegraphics[width=\linewidth]{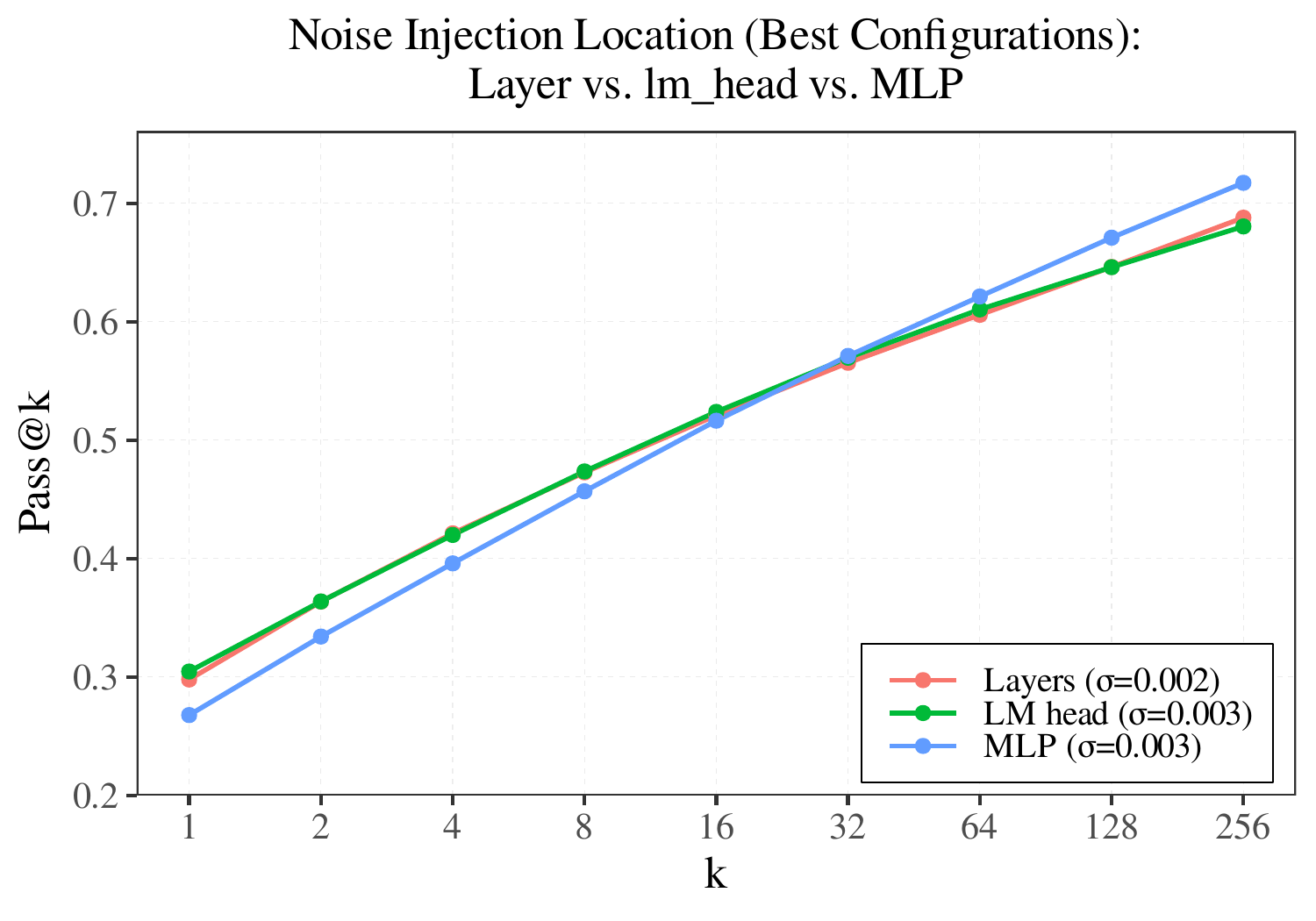}
    \caption{Noise injection location (best settings). Average pass@k  for the best noise scale at each injection place (Whole Layers, \texttt{lm\_head}, and MLP). MLP injection attains the largest gains at high $k$.}
    \label{fig:where_inject_noise_main}
\end{figure}
Theoretical work by Plappert et al.~\cite{plappert2018parameter} suggests that parameter space noise yields optimal performance  when using normalization  between perturbed layers. In standard Transformer architectures, \ul{the MLP blocks naturally satisfy this structural criterion}. To empirically validate this hypothesis and demonstrate the method's generalization beyond specific math-tuned models, we conducted ablation studies using the general-purpose \textbf{Qwen2.5-7B} model. We compared noise injection across three distinct location: full decoding layers, the language modeling head (\texttt{lm\_head}), and the MLP blocks. As illustrated in Figure~\ref{fig:where_inject_noise_main}, injecting noise exclusively into the MLP layers yields a superior reasoning capability boundary, demonstrating significantly higher pass@$k$ performance at large sampling budgets ($k \ge 128$) compared to other strategies. Comprehensive results are provided in Appendix Figure~\ref{fig:where_inject_noise_detail}. \paragraph{Additional evidence on \textbf{Qwen3-4B-Base}.}
We further evaluate PSN on \textbf{Qwen3-4B-Base} using our default PSN setting (MLP-only perturbation). Table~\ref{tab:q2_qwen3_4bbase} shows the same exploration--exploitation pattern observed in larger models: PSN slightly decreases low-budget reliability (pass@1/2) but expands the high-budget boundary (pass@256). Notably, on harder benchmarks PSN improves pass@256 on AIME24 by \textbf{+2.7pp} (62.8$\rightarrow$65.5) and on AIME25 by \textbf{+3.7pp} (58.9$\rightarrow$62.6), consistent with PSN unlocking additional reasoning modes under large sampling. Detailed per-benchmark results for Qwen3-4B-Base are reported in Appendix Table~\ref{tab:q2_qwen3_4bbase_detail}.

\begin{table}[t]
\centering
\setlength{\tabcolsep}{6pt}
\resizebox{\linewidth}{!}{
\begin{tabular}{lccc}
\toprule
\textbf{Method (Qwen3-4B-Base)} & \textbf{Avg pass@1} & \textbf{Avg pass@2} & \textbf{Avg pass@256} \\
\midrule
Clean ($\sigma=0$) & 33.8\% & 41.5\% & 72.0\% \\
PSN ($\sigma=0.001$) & 31.9\% & 40.5\% & \textbf{72.5\%} \\
\bottomrule
\end{tabular}
}
\caption{\textbf{Qwen3-4B-Base generalization.} Average pass@$k$ (\%) over five math benchmarks (AIME24/25, AMC23, Minerva-Math, OlympiadBench) comparing clean decoding vs.\ PSN. PSN expands the high-budget reasoning boundary (avg pass@256 +0.5pp) while slightly reducing low-budget reliability, matching the exploration--exploitation trade-off.}
\label{tab:q2_qwen3_4bbase}
\end{table}


\ul{\textbf{Q3: Is Parameter space noise more effective than action-space noise with respect to long-trajectory(CoT) reasoning in RLVR method?}}

\label{q:action_noise_scaling}
\begin{tcolorbox}[
    colback=blue!5!white,  
    colframe=blue!5!white, 
    boxsep=1pt,            
    left=2pt,              
    right=2pt,             
    top=2pt,               
    bottom=2pt,            
    arc=2pt                
]
A: Yes, parameter-space noise outperforms both \textcolor{blue}{training} and \textcolor{orange}{evaluation} time action-space noise by preserving long-trajectory reasoning (CoT) consistency.
\end{tcolorbox}

\begin{figure}[]
    \centering
    \includegraphics[width=\linewidth]{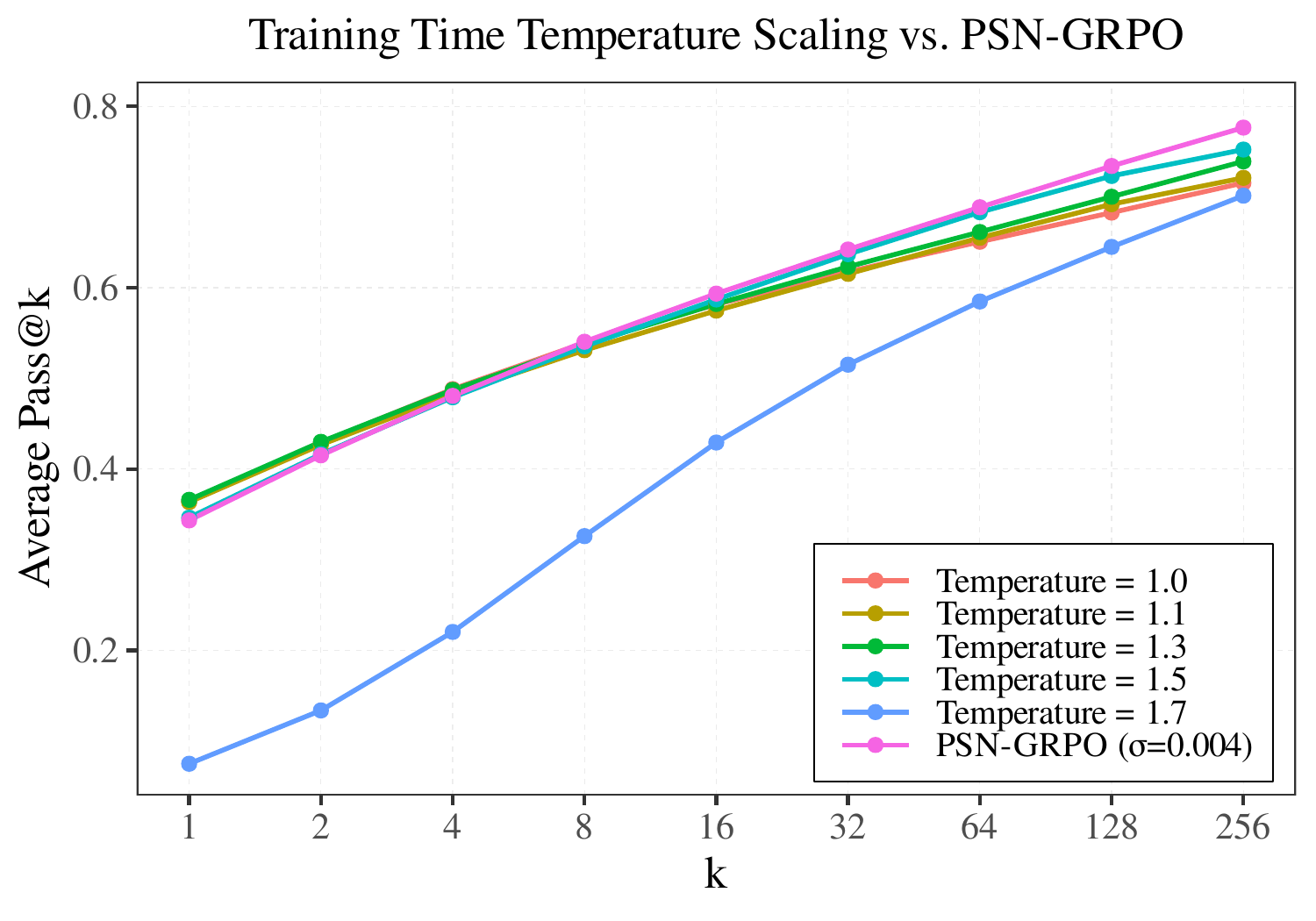}
    \caption{Average performance across benchmarks with different training time temperature. Increasing temperature beyond $1.5$ degrades  overall performance. }
    \label{fig:training_temp_scaling}
\end{figure}

We evaluate the efficacy of parameter-space exploration against action-space noise baselines. First, we compare PSN-GRPO against \textcolor{blue}{training-time action-space noise}, implemented via temperature scaling ($T \in \{1.0, \dots, 1.7\}$) ~\cite{du2025optimizingtemperaturelanguagemodels}. As shown in Figure~\ref{fig:training_temp_scaling}, increasing temperature beyond $1.5$ degrades performance. We attribute this failure mode to the local, unstructured nature of token-level perturbations: because action-space noise is typically uncorrelated across time steps, it accumulates into ``logical drift'' that disrupts the global coherence required for long-horizon Chain-of-Thought (CoT) reasoning.

In contrast, PSN perturbs the policy parameters prior to generation, inducing \textit{trajectory-level consistency}. Consequently, PSN-GRPO demonstrates superior scaling behavior, with performance benefits becoming increasingly \ul{pronounced as the length of reasoning trajectories grow}, as shown in Table~\ref{tab:merged_psn_advantages}, and  detailed results are shown in Appendix Figure~\ref{fig:training_temp_scaling_detail}. Notably, while the performance gain over the baseline is marginal ($<1\%$) on the shorter AMC 23 dataset (average response length is 738 tokens), the gap widens to \textbf{$+8.9\%$} (pass@256) on the hard task, AIME 24 benchmark (average response length is 1,978 tokens). Furthermore, PSN-GRPO consistently exceeds the performance ceiling of \textcolor{orange}{evaluation-time temperature scaling} (typically $T=1.5$), which requires expensive tuning and lacks the coherent exploration necessary for difficult tasks as shown in Table~\ref{tab:merged_psn_advantages}, and  detailed results are shown in  Appendix Figure~\ref{fig:evaluation_noise_scaling_detail}.
\\
\\

\begin{table}[t]
\centering

\setlength{\tabcolsep}{5pt} 
\resizebox{\linewidth}{!}{
\begin{tabular}{l c l cc cc cc}
\toprule
\multirow{2}{*}{\textbf{Dataset}} & \multirow{2}{*}{\makecell[c]{\textbf{Avg. Len} \\ \textbf{(PSN)}}} & \multirow{2}{*}{\textbf{Method}} & \multicolumn{2}{c}{\textbf{pass@64}} & \multicolumn{2}{c}{\textbf{pass@128}} & \multicolumn{2}{c}{\textbf{pass@256}} \\
\cmidrule(lr){4-5} \cmidrule(lr){6-7} \cmidrule(lr){8-9}
& & & \textbf{Score} & \textbf{Gap} & \textbf{Score} & \textbf{Gap} & \textbf{Score} & \textbf{Gap} \\
\midrule
\multirow{3}{*}{AMC 23} & \multirow{3}{*}{ \textcolor{darkblue1}{738} } & PSN-GRPO & \underline{\textbf{96.9\%}} & - & \underline{\textbf{99.3\%}} & - & \underline{\textbf{100.0\%}} & - \\
& & B.Train & 95.7\% & \textcolor{darkgreen}{+1.2\%} & 98.5\% & \textcolor{darkgreen}{+0.8\%} & 99.9\% & \textcolor{darkgreen}{+0.1\%} \\
& & B.Eval & 94.6\% & \textcolor{darkgreen}{+2.3\%} & 97.0\% & \textcolor{darkgreen}{+2.3\%} & 99.6\% & \textcolor{darkgreen}{+0.4\%} \\
\midrule
\multirow{3}{*}{Olympiad} & \multirow{3}{*}{\textcolor{darkblue2}{923}} & PSN-GRPO & \underline{\textbf{73.5\%}} & - & \underline{\textbf{77.0\%}} & - & \underline{\textbf{80.1\%}} & - \\
& & B.Train & 72.0\% & \textcolor{darkgreen}{+1.5\%} & 75.0\% & \textcolor{darkgreen}{+2.0\%} & 77.9\% & \textcolor{darkgreen}{+2.2\%} \\
& & B.Eval & 72.6\% & \textcolor{darkgreen}{+0.9\%} & 75.8\% & \textcolor{darkgreen}{+1.2\%} & 78.7\% & \textcolor{darkgreen}{+1.3\%} \\
\midrule
\multirow{3}{*}{AIME 25} & \multirow{3}{*}{\textcolor{darkblue3}{1030}} & PSN-GRPO & \underline{\textbf{49.9\%}} & - & \underline{\textbf{56.1\%}} & - & \underline{\textbf{62.2\%}} & - \\
& & B.Train & 47.3\% & \textcolor{darkgreen}{+2.6\%} & 54.1\% & \textcolor{darkgreen}{+2.0\%} & 61.6\% & \textcolor{darkgreen}{+0.6\%} \\
& & B.Eval & 48.6\% & \textcolor{darkgreen}{+1.3\%} & 53.9\% & \textcolor{darkgreen}{+2.2\%} & 61.3\% & \textcolor{darkgreen}{+0.9\%} \\
\midrule
\multirow{3}{*}{AIME 24} & \multirow{3}{*}{\textcolor{darkblue4}{\textbf{1978}}} & PSN-GRPO & \underline{\textbf{65.2\%}} & - & \underline{\textbf{73.0\%}} & - & \underline{\textbf{81.6\%}} & - \\
& & B.Train & 64.6\% & \textcolor{darkgreen1}{+0.6\%} & 69.0\% & \textcolor{darkgreen3}{+4.0\%} & 72.7\% & \textcolor{darkgreen4}{+8.9\%} \\
& & B.Eval & 58.9\% & \textcolor{darkgreen4}{+6.3\%} & 64.7\% & \textcolor{darkgreen4}{+8.3\%} & 71.7\% & \textcolor{darkgreen4}{+9.9\%} \\
\bottomrule
\end{tabular}
}
\caption{\textbf{Performance comparison across mathematical benchmarks.} We report pass@$k$ scores (\%) for PSN-GRPO ($\sigma=0.004$) relative to the \textcolor{blue}{best training-time} (B.Train) and \textcolor{orange}{best evaluation-time} (B.Eval) temperature-scaling baselines. \textbf{Gap} denotes the absolute percentage point improvement of PSN-GRPO over the respective baseline. \textbf{Key Finding:} PSN-GRPO outperforms action-space noise by inducing \textit{trajectory-level consistency}, effectively mitigating the logical drift observed in temperature scaling—a benefit that becomes increasingly pronounced on long-horizon tasks (e.g., AIME 24 with average response length around 2k).}
\label{tab:merged_psn_advantages}
\end{table}

\ul{\textbf{Q4: Is Truncated Importance Sampling (TIS) necessary?}}
\begin{tcolorbox}[
    colback=blue!5!white,  
    colframe=blue!5!white, 
    boxsep=1pt,            
    left=2pt,              
    right=2pt,             
    top=2pt,               
    bottom=2pt,            
    arc=2pt                
]
A: Yes, TIS is crucial for stabilizing training by mitigating the off-policy mismatch.
\end{tcolorbox}
Since PSN-RLVR generates rollouts using a perturbed policy $\pi_{noisy}$ but updates a clean policy $\pi_{clean}$, a distribution mismatch arises. Table~\ref{tab:q4_tis} compares standard GRPO and PSN-GRPO with and without TIS correction. We observe that while TIS provides negligible benefits to standard GRPO (where the sampling and training policies are identical), it significantly boosts the performance of PSN-GRPO (e.g., increasing pass@256 from $74.33\%$ to $76.94\%$). This confirms that properly handling the importance ratio clipping is essential when leveraging exploratory data from parameter-perturbed policies.
\begin{table}[h]
\centering
\resizebox{\linewidth}{!}{
    \begin{tabular}{l|c c c}
    \hline
    \diagbox{Method Name}{Avg\\ of Datasets} & pass@1 & pass@128 & pass@256 \\ \hline
        PSN-GRPO & 36.01\% &  71.10\%&  74.33\%\\ \hline
    PSN-GRPO\_with\_TIS & \textbf{36.15\%} & \textbf{73.07\%} & \textbf{76.94\%} \\ \hline
    \end{tabular}
} 
\caption{\textbf{Impact of Truncated Importance Sampling (TIS).} The results show that TIS correction significantly boosts the performance of PSN-GRPO (e.g., increasing pass@256 from 74.33\% to 76.94\%) by effectively mitigating the off-policy mismatch between the perturbed sampling policy and the clean training policy.}\label{tab:q4_tis}
\end{table}

\begin{figure}[t]
    \centering
    \includegraphics[width=\linewidth]{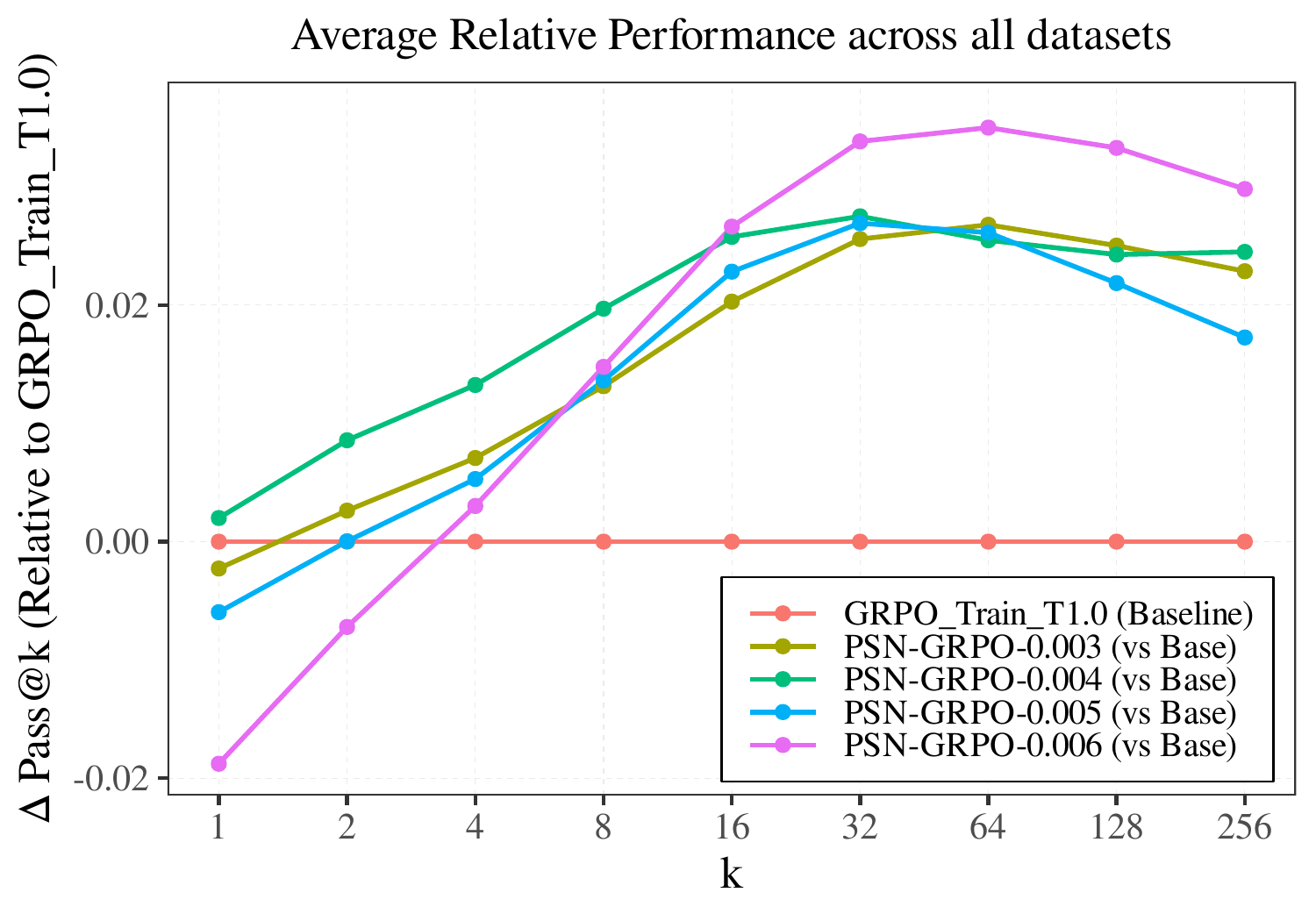}
\caption{\textbf{Impact of Noise Magnitude $\sigma$ on Exploration-Exploitation.} Increasing noise scale (e.g., $\sigma=0.006$) maximizes the reasoning capability boundary at high $k$ but degrades pass@1 performance. Moderate scales ($\sigma \in \{0.004, 0.005\}$) provide the optimal balance between exploitation reliability and exploratory diversity across benchmarks.}    \label{fig:q5_noise_scaling}
\end{figure}

\ul{\textbf{Q5: How does performance scale with noise magnitude?}}
\begin{tcolorbox}[
    colback=blue!5!white,  
    colframe=blue!5!white, 
    boxsep=1pt,            
    left=2pt,              
    right=2pt,             
    top=2pt,               
    bottom=2pt,            
    arc=2pt                
]
A: A moderate noise level (e.g., $\sigma \in \{0.004, 0.005\}$) tends to achieve the optimal exploration-exploitation balance. While larger noise (e.g., $\sigma=0.006$) improves pass@k for larger $k$ (e.g., $k=128, 256$), it detrimentally affects performance at lower pass@k.
\end{tcolorbox}
Figure~\ref{fig:q5_noise_scaling} illustrates the trade-off inherent in noise magnitude selection. We find that larger noise scales (e.g., $\sigma=0.006$) yield the highest pass@256 scores, indicating successful exploration of the outer reaches of the solution space. However, this comes at the cost of lower pass@1 performance compared to moderate noise levels (e.g., $\sigma \in \{0.004, 0.005\}$). For general applications, a moderate noise level provides the optimal balance between exploitation (reliability) and exploration (diversity).

\ul{\textbf{Q6: Is adaptive noise scheduling better than fixed noise?}}
\begin{tcolorbox}[
    colback=blue!5!white,  
    colframe=blue!5!white, 
    boxsep=1pt,            
    left=2pt,              
    right=2pt,             
    top=2pt,               
    bottom=2pt,            
    arc=2pt                
]
A: Yes,  our proposed real-time certainty-aware scheduling achieves a superior exploration-stability trade-off.
\end{tcolorbox}
We benchmark our compute-aware adaptive noise schedules (Variant I and Variant II, as detailed in Section~\ref{subsec:adaptive_noise}) against fixed noise schedules and the GRPO baseline. The adaptive approach modulates $\sigma$ based on semantic similarity and self-certainty signals (Equation~\ref{eq:real_time_update}).
Our results highlight a critical trade-off:  as shown in Table~\ref{tab:adaptive_noise_advantage}, and detailed result in Table~\ref{tab:adaptive_noise_advantage_detail}. \textbf{Variant I} (non-real-time), despite incurring small inference overhead, suffers from feedback lag that degrades performance relative to standard GRPO. Conversely, \textbf{Variant II} (real-time), while introducing a marginal computational cost, effectively synchronizes noise scaling with the model's current state. This results in significant gains in both sample efficiency (e.g., pass@2) and the reasoning capability boundary (e.g., pass@256), consistently outperforming both fixed-noise baselines and naive GRPO.
\begin{table}[t]
\centering
\setlength{\tabcolsep}{5pt} 
\resizebox{\linewidth}{!}{
\begin{tabular}{l l cc cc cc}
\toprule
\multirow{2}{*}{\textbf{Dataset}} & \multirow{2}{*}{\textbf{Method}} & \multicolumn{2}{c}{\textbf{pass@2}} & \multicolumn{2}{c}{\textbf{pass@4}} & \multicolumn{2}{c}{\textbf{pass@256}} \\
\cmidrule(lr){3-4} \cmidrule(lr){5-6} \cmidrule(lr){7-8}
& & \textbf{Score} & \textbf{Gap} & \textbf{Score} & \textbf{Gap} & \textbf{Score} & \textbf{Gap} \\
\midrule
\multirow{4}{*}{Average} & PSN Var-II & \underline{\textbf{44.1\%}} & - & \underline{\textbf{50.6\%}} & - & \underline{\textbf{79.5\%}} & - \\
& GRPO Train & 42.7\% & \textcolor{darkgreen}{+1.3\%} & 48.7\% & \textcolor{darkgreen}{+1.9\%} & 74.7\% & \textcolor{darkgreen4}{+4.8\%} \\
& PSN Fixed & 43.6\% & \textcolor{darkgreen}{+0.5\%} & 50.0\% & \textcolor{darkgreen}{+0.6\%} & 77.1\% & \textcolor{darkgreen}{+2.4\%} \\
& PSN Var-I & 42.5\% & \textcolor{darkgreen}{+1.5\%} & 48.7\% & \textcolor{darkgreen2}{+1.9\%} & 75.1\% & \textcolor{darkgreen4}{+4.4\%} \\
\bottomrule
\end{tabular}
}
\caption{\textbf{Performance comparison across mathematical benchmarks.} We report pass@$k$ scores (\%) for our adaptive variants (PSN Var-I/II) against the standard GRPO baseline and Fixed-Noise PSN ($\sigma=0.004$). \textbf{Gap} denotes the absolute percentage point improvement of \textit{PSN Var-II} over the compared method. Notably, the non-real-time schedule (Var-I) hurts performance due to feedback latency. In contrast, the proposed lightweight real-time schedule (Var-II) yields the best performance, improving both sample efficiency (e.g., +1.3\% pass@2 vs. GRPO) and the reasoning capability boundary (e.g., +4.8\% pass@256 vs. GRPO).}
\label{tab:adaptive_noise_advantage}
\end{table}

\ul{\textbf{Q7: How does PSN-RLVR compared with  other exploration methods and Is it orthogonal to these method?}}
\label{sec:res_q7_orthogonal}
\begin{tcolorbox}[
    colback=blue!5!white,  
    colframe=blue!5!white, 
    boxsep=1pt,            
    left=2pt,              
    right=2pt,             
    top=2pt,               
    bottom=2pt,            
    arc=2pt                
]
A: Yes, PSN-RLVR outperforms other mainstream exploration baselines  and is orthogonal to these methods, yielding additive performance gains when combined
\end{tcolorbox}

First, we compare PSN-RLVR against  two mainstreams method that increase explore capability pass@K~\cite{Chen2025PasskTF} and RLVR Decomposed~\cite{zhu2025surprisingeffectivenessnegativereinforcement} . As illustrated in Figure~\ref{fig:different_methods}, PSN-RLVR not only achieves a superior reasoning capability boundary (higher pass@K at large budgets) but also exhibits significantly higher semantic and operation diversity compared to the baselines. This suggests that parameter-space perturbations induce novel reasoning modes that objective or data modifications alone fail to uncover. Second, we demonstrate that PSN acts as a complementary exploration mechanism. As shown in Table~\ref{tab:orthogonal}, combining PSN with pass@K training (PSN-GRPO-pass@K) further boosts performance, improving the average pass@256 from 76.37\% to 79.12\% over the strong GRPO-pass@K baseline. This result confirms that PSN is orthogonal to other strategies and can be effectively composed with them to maximize exploration.\\
\begin{table}[h]
\centering
\resizebox{\linewidth}{!}{
    \begin{tabular}{l|c c c}
    \hline
    \diagbox{Method Name}{Avg\\ of Datasets} & pass@64 & pass@128 & pass@256 \\ \hline
    GRPO-pass@K & 69.77\% & 73.48\% & 76.37\% \\ \hline
    PSN-GRPO-pass@K & \textbf{70.35\%} & \textbf{74.82\%} & \textbf{79.12\%} \\ \hline
    \end{tabular}
} 
\caption{\textbf{PSN-GRPO is complementary to Pass@k training.} Combining Pass@k-based training with PSN-GRPO yields higher performance than either component alone.}
\label{tab:orthogonal}
\end{table}\\
\paragraph{Finding: PSN-RLVR discovers qualitatively new solution strategies.}
To probe whether the gains in \emph{reasoning capability boundary} reflect genuine exploration (rather than mere reweighting), we conduct an Gemini-assisted~\cite{Comanici2025Gemini2P} qualitative analysis on AIME~2024 under our standard evaluation protocol (30 problems; $n{=}300$ rollouts per problem).
We focus on the subset of problems where the \emph{base/original} model fails on all $300$ rollouts, yet PSN-RLVR produces at least one correct rollout.
Across these cases, we find that the successful PSN-RLVR traces typically employ solution perspective that are absent from the base model’s rollout set, indicating that PSN-RLVR can access \emph{new reasoning modes} rather than only improving selection among pre-existing trajectories. Representative examples are provided in Appendix~\ref{sec:detailed_new_solution}.
\vspace{-1em}
\section{Limitations}
PSN-RLVR is most effective for long-horizon reasoning tasks requiring global consistency. In simpler, short-sequence tasks where token-level stochasticity suffices, PSN may yield diminishing returns. This limitation is mitigated by our real-time adaptive scheduler (Table~\ref{tab:adaptive_noise_advantage_detail}), though further investigation is warranted.

\section{Conclusion}
We introduced PSN-RLVR, the first systematic study of parameter-space noise for RLVR-trained language models, motivated by the observation that standard RLVR can saturate, primarily by  improving selection among already-likely trajectories. By perturbing parameters (rather than tokens) to produce rollout-consistent exploration, PSN-RLVR improves long-horizon reasoning and yields larger gains under high sampling budgets. We showed that injecting noise into Transformer MLP/FFN blocks offers a favorable stability--exploration trade-off, and that truncated importance sampling is essential to reliably exploit exploratory data from perturbed policies. Finally, our lightweight certainty- and semantics-aware scheduling achieves robust performance without additional rollout overhead, and PSN composes with existing RLVR exploration strategies to further extend the achievable pass@k frontier.

\section*{Impact Statement}
This work advances RLVR for reasoning by introducing a practical exploration mechanism---parameter-space noise with stable off-policy correction and compute-aware scheduling---that improves high-budget sampling performance and diversity on verifiable math benchmarks. The primary positive impact is enabling more reliable and efficient discovery of correct solution strategies in domains with automated checkers (e.g., education and software tooling). As with other stronger reasoning models, misuse risks include facilitating automated generation of deceptive content or enabling harmful applications; our method does not inherently mitigate these risks. We recommend standard safeguards for model deployment, including controlled access, monitoring, and domain-specific evaluation beyond verifiable tasks.

\nocite{langley00}

\bibliography{example_paper}

@inproceedings{langley00,
 author    = {P. Langley},
 title     = {Crafting Papers on Machine Learning},
 year      = {2000},
 pages     = {1207--1216},
 editor    = {Pat Langley},
 booktitle     = {Proceedings of the 17th International Conference
              on Machine Learning (ICML 2000)},
 address   = {Stanford, CA},
 publisher = {Morgan Kaufmann}
}

@misc{wu2026invisibleleashrlvrescape,
      title={The Invisible Leash: Why RLVR May or May Not Escape Its Origin}, 
      author={Fang Wu and Weihao Xuan and Ximing Lu and Mingjie Liu and Yi Dong and Zaid Harchaoui and Yejin Choi},
      year={2026},
      eprint={2507.14843},
      archivePrefix={arXiv},
      primaryClass={cs.LG},
      url={https://arxiv.org/abs/2507.14843}, 
}

@misc{steinfeldt2024bert_mlm_arxiv_mp_class_zbmath_model,
  author       = {Steinfeldt, Christian and Mihaljevic, Helena},
  title        = {{Bert-MLM\_arXiv-MP-class\_zbMath}: A sentence-transformers model for similarity of short mathematical texts},
  howpublished = {\url{https://huggingface.co/math-similarity/Bert-MLM_arXiv-MP-class_zbMath}},
  year         = {2024},
  note         = {Hugging Face model card. Accessed: 2026-01-25}
}

@article{sheng2024hybridflow,
  title   = {HybridFlow: A Flexible and Efficient RLHF Framework},
  author  = {Guangming Sheng and Chi Zhang and Zilingfeng Ye and Xibin Wu and Wang Zhang and Ru Zhang and Yanghua Peng and Haibin Lin and Chuan Wu},
  year    = {2024},
  journal = {arXiv preprint arXiv: 2409.19256}
}

@misc{lambert2025tulu3pushingfrontiers,
      title={Tulu 3: Pushing Frontiers in Open Language Model Post-Training}, 
      author={Nathan Lambert and Jacob Morrison and Valentina Pyatkin and Shengyi Huang and Hamish Ivison and Faeze Brahman and Lester James V. Miranda and Alisa Liu and Nouha Dziri and Shane Lyu and Yuling Gu and Saumya Malik and Victoria Graf and Jena D. Hwang and Jiangjiang Yang and Ronan Le Bras and Oyvind Tafjord and Chris Wilhelm and Luca Soldaini and Noah A. Smith and Yizhong Wang and Pradeep Dasigi and Hannaneh Hajishirzi},
      year={2025},
      eprint={2411.15124},
      archivePrefix={arXiv},
      primaryClass={cs.CL},
      url={https://arxiv.org/abs/2411.15124}, 
}

@article{Comanici2025Gemini2P,
  title={Gemini 2.5: Pushing the Frontier with Advanced Reasoning, Multimodality, Long Context, and Next Generation Agentic Capabilities},
  author={Gemini Team},
  journal={ArXiv},
  year={2025},
  volume={abs/2507.06261},
  url={https://api.semanticscholar.org/CorpusID:280151524}
}

@misc{aime24,
      title={AIME. AIME problems and solutions,}, 
      year={2025},
      url={https://artofproblemsolving.com/wiki/index.php/AIME_Problems_and_Solutions}, 
}

@misc{cheng2025fastllmposttrainingdecoupled,
      title={Fast LLM Post-training via Decoupled and Fastest-of-N Speculation}, 
      author={Rongxin Cheng and Kai Zhou and Xingda Wei and Siyuan Liu and Mingcong Han and Mingjing Ai and Yeju Zhou and Baoquan Zhong and Wencong Xiao and Rong Chen and Haibo Chen},
      year={2025},
      eprint={2511.16193},
      archivePrefix={arXiv},
      primaryClass={cs.DC},
      url={https://arxiv.org/abs/2511.16193}, 
}

@article{hendrycksmath2021,
    title={Measuring Mathematical Problem Solving With the MATH Dataset},
    author={Dan Hendrycks
    and Collin Burns
    and Saurav Kadavath
    and Akul Arora
    and Steven Basart
    and Eric Tang
    and Dawn Song
    and Jacob Steinhardt},
    journal={arXiv preprint arXiv:2103.03874},
    year={2021}
}

@misc{gsm8k,
      title={Training Verifiers to Solve Math Word Problems}, 
      author={Karl Cobbe and Vineet Kosaraju and Mohammad Bavarian and Mark Chen and Heewoo Jun and Lukasz Kaiser and Matthias Plappert and Jerry Tworek and Jacob Hilton and Reiichiro Nakano and Christopher Hesse and John Schulman},
      year={2021},
      eprint={2110.14168},
      archivePrefix={arXiv},
      primaryClass={cs.LG},
      url={https://arxiv.org/abs/2110.14168}, 
}

@misc{numina_math_datasets,
  author = {Jia LI and Edward Beeching and Lewis Tunstall and Ben Lipkin and Roman Soletskyi and Shengyi Costa Huang and Kashif Rasul and Longhui Yu and Albert Jiang and Ziju Shen and Zihan Qin and Bin Dong and Li Zhou and Yann Fleureau and Guillaume Lample and Stanislas Polu},
  title = {NuminaMath},
  year = {2024},
  publisher = {Numina},
  journal = {Hugging Face repository},
  howpublished = {\url{[https://huggingface.co/AI-MO/NuminaMath-CoT](https://github.com/project-numina/aimo-progress-prize/blob/main/report/numina_dataset.pdf)}}
}

@misc{zeng2025simplerlzooinvestigatingtamingzero,
      title={SimpleRL-Zoo: Investigating and Taming Zero Reinforcement Learning for Open Base Models in the Wild}, 
      author={Weihao Zeng and Yuzhen Huang and Qian Liu and Wei Liu and Keqing He and Zejun Ma and Junxian He},
      year={2025},
      eprint={2503.18892},
      archivePrefix={arXiv},
      primaryClass={cs.LG},
      url={https://arxiv.org/abs/2503.18892}, 
}

@misc{qwen2025qwen25technicalreport,
      title={Qwen2.5 Technical Report}, 
      author={Qwen Team},
      year={2025},
      eprint={2412.15115},
      archivePrefix={arXiv},
      primaryClass={cs.CL},
      url={https://arxiv.org/abs/2412.15115}, 
}

@misc{yang2024qwen25mathtechnicalreportmathematical,
      title={Qwen2.5-Math Technical Report: Toward Mathematical Expert Model via Self-Improvement}, 
      author={An Yang and Beichen Zhang and Binyuan Hui and Bofei Gao and Bowen Yu and Chengpeng Li and Dayiheng Liu and Jianhong Tu and Jingren Zhou and Junyang Lin and Keming Lu and Mingfeng Xue and Runji Lin and Tianyu Liu and Xingzhang Ren and Zhenru Zhang},
      year={2024},
      eprint={2409.12122},
      archivePrefix={arXiv},
      primaryClass={cs.CL},
      url={https://arxiv.org/abs/2409.12122}, 
}

@misc{fang2025wrongperplexitylongcontextlanguage,
      title={What is Wrong with Perplexity for Long-context Language Modeling?}, 
      author={Lizhe Fang and Yifei Wang and Zhaoyang Liu and Chenheng Zhang and Stefanie Jegelka and Jinyang Gao and Bolin Ding and Yisen Wang},
      year={2025},
      eprint={2410.23771},
      archivePrefix={arXiv},
      primaryClass={cs.CL},
      url={https://arxiv.org/abs/2410.23771}, 
}

@misc{kang2025scalablebestofnselectionlarge,
      title={Scalable Best-of-N Selection for Large Language Models via Self-Certainty}, 
      author={Zhewei Kang and Xuandong Zhao and Dawn Song},
      year={2025},
      eprint={2502.18581},
      archivePrefix={arXiv},
      primaryClass={cs.CL},
      url={https://arxiv.org/abs/2502.18581}, 
}

@article{Zhan2025MindYE,
  title={Mind Your Entropy: From Maximum Entropy to Trajectory Entropy-Constrained RL},
  author={Guojian Zhan and Likun Wang and Pengcheng Wang and Feihong Zhang and Jingliang Duan and Masayoshi Tomizuka and Shengbo Eben Li},
  journal={ArXiv},
  year={2025},
  volume={abs/2511.11592},
  url={https://api.semanticscholar.org/CorpusID:283071650}
}

@misc{yue2025does,
      title={Does Reinforcement Learning Really Incentivize Reasoning Capacity in LLMs Beyond the Base Model?}, 
      author={Yang Yue and Zhiqi Chen and Rui Lu and Andrew Zhao and Zhaokai Wang and Yang Yue and Shiji Song and Gao Huang},
      year={2025},
      eprint={2504.13837},
      archivePrefix={arXiv},
      primaryClass={cs.AI},
      url={https://arxiv.org/abs/2504.13837}, 
}

@misc{wang2024mathvista,
      title={MathVista: Evaluating Mathematical Reasoning of Foundation Models in Visual Contexts}, 
      author={Pan Lu and Hritik Bansal and Tony Xia and Jiacheng Liu and Chunyuan Li and Hannaneh Hajishirzi and Hao Cheng and Kai-Wei Chang and Michel Galley and Jianfeng Gao},
      year={2024},
      eprint={2310.02255},
      archivePrefix={arXiv},
      primaryClass={cs.CV},
      url={https://arxiv.org/abs/2310.02255}, 
}

@article{ouyang2022training,
  title={Training language models to follow instructions with human feedback},
  author={Ouyang, Long and others},
  journal={Advances in Neural Information Processing Systems},
  volume={35},
  pages={27730--27744},
  year={2022}
}

@misc{schulman2017proximalpolicyoptimizationalgorithms,
      title={Proximal Policy Optimization Algorithms}, 
      author={John Schulman and Filip Wolski and Prafulla Dhariwal and Alec Radford and Oleg Klimov},
      year={2017},
      eprint={1707.06347},
      archivePrefix={arXiv},
      primaryClass={cs.LG},
      url={https://arxiv.org/abs/1707.06347}, 
}

@misc{he2025historyrhymesacceleratingllm,
      title={History Rhymes: Accelerating LLM Reinforcement Learning with RhymeRL}, 
      author={Jingkai He and Tianjian Li and Erhu Feng and Dong Du and Qian Liu and Tao Liu and Yubin Xia and Haibo Chen},
      year={2025},
      eprint={2508.18588},
      archivePrefix={arXiv},
      primaryClass={cs.LG},
      url={https://arxiv.org/abs/2508.18588}, 
}

@inproceedings{fortunato2018noisy,
  title        = {Noisy Networks for Exploration},
  author       = {Fortunato, Meire and Azar, Mohammad Gheshlaghi and Piot, Bilal and Menick, Jacob and Osband, Ian and Graves, Alex and Mnih, Vlad and Munos, R{\'e}mi and Hassabis, Demis and Legg, Shane and others},
  booktitle    = {International Conference on Learning Representations (ICLR)},
  year         = {2018},
  note         = {arXiv:1706.10295}
}

@misc{zhu2025surprisingeffectivenessnegativereinforcement,
      title={The Surprising Effectiveness of Negative Reinforcement in LLM Reasoning}, 
      author={Xinyu Zhu and Mengzhou Xia and Zhepei Wei and Wei-Lin Chen and Danqi Chen and Yu Meng},
      year={2025},
      eprint={2506.01347},
      archivePrefix={arXiv},
      primaryClass={cs.CL},
      url={https://arxiv.org/abs/2506.01347}, 
}

@misc{jiang2025selectiveexpertguidanceeffective,
      title={Selective Expert Guidance for Effective and Diverse Exploration in Reinforcement Learning of LLMs}, 
      author={Zishang Jiang and Jinyi Han and Tingyun Li and Xinyi Wang and Sihang Jiang and Jiaqing Liang and Zhaoqian Dai and Shuguang Ma and Fei Yu and Yanghua Xiao},
      year={2025},
      eprint={2510.04140},
      archivePrefix={arXiv},
      primaryClass={cs.AI},
      url={https://arxiv.org/abs/2510.04140}, 
}

@misc{li2025questaexpandingreasoningcapacity,
      title={QuestA: Expanding Reasoning Capacity in LLMs via Question Augmentation}, 
      author={Jiazheng Li and Hongzhou Lin and Hong Lu and Kaiyue Wen and Zaiwen Yang and Jiaxuan Gao and Yi Wu and Jingzhao Zhang},
      year={2025},
      eprint={2507.13266},
      archivePrefix={arXiv},
      primaryClass={cs.CL},
      url={https://arxiv.org/abs/2507.13266}, 
}

@misc{dong2025rlpluscounteringcapabilityboundary,
      title={RL-PLUS: Countering Capability Boundary Collapse of LLMs in Reinforcement Learning with Hybrid-policy Optimization}, 
      author={Yihong Dong and Xue Jiang and Yongding Tao and Huanyu Liu and Kechi Zhang and Lili Mou and Rongyu Cao and Yingwei Ma and Jue Chen and Binhua Li and Zhi Jin and Fei Huang and Yongbin Li and Ge Li},
      year={2025},
      eprint={2508.00222},
      archivePrefix={arXiv},
      primaryClass={cs.AI},
      url={https://arxiv.org/abs/2508.00222}, 
}

@misc{liang2025pass1selfplayvariationalproblem,
      title={Beyond Pass@1: Self-Play with Variational Problem Synthesis Sustains RLVR}, 
      author={Xiao Liang and Zhongzhi Li and Yeyun Gong and Yelong Shen and Ying Nian Wu and Zhijiang Guo and Weizhu Chen},
      year={2025},
      eprint={2508.14029},
      archivePrefix={arXiv},
      primaryClass={cs.CL},
      url={https://arxiv.org/abs/2508.14029}, 
}

@misc{cui2025entropymechanismreinforcementlearning,
      title={The Entropy Mechanism of Reinforcement Learning for Reasoning Language Models}, 
      author={Ganqu Cui and Yuchen Zhang and Jiacheng Chen and Lifan Yuan and Zhi Wang and Yuxin Zuo and Haozhan Li and Yuchen Fan and Huayu Chen and Weize Chen and Zhiyuan Liu and Hao Peng and Lei Bai and Wanli Ouyang and Yu Cheng and Bowen Zhou and Ning Ding},
      year={2025},
      eprint={2505.22617},
      archivePrefix={arXiv},
      primaryClass={cs.LG},
      url={https://arxiv.org/abs/2505.22617}, 
}

@misc{cheng2025reasoningexplorationentropyperspective,
      title={Reasoning with Exploration: An Entropy Perspective}, 
      author={Daixuan Cheng and Shaohan Huang and Xuekai Zhu and Bo Dai and Wayne Xin Zhao and Zhenliang Zhang and Furu Wei},
      year={2025},
      eprint={2506.14758},
      archivePrefix={arXiv},
      primaryClass={cs.CL},
      url={https://arxiv.org/abs/2506.14758}, 
}

@misc{peng2025simkosimplepasskpolicy,
      title={SimKO: Simple Pass@K Policy Optimization}, 
      author={Ruotian Peng and Yi Ren and Zhouliang Yu and Weiyang Liu and Yandong Wen},
      year={2025},
      eprint={2510.14807},
      archivePrefix={arXiv},
      primaryClass={cs.AI},
      url={https://arxiv.org/abs/2510.14807}, 
}

@article{Chen2025PasskTF,
  title={Pass@k Training for Adaptively Balancing Exploration and Exploitation of Large Reasoning Models},
  author={Zhipeng Chen and Xiaobo Qin and Youbin Wu and Yue Ling and Qinghao Ye and Wayne Xin Zhao and Guang Shi},
  journal={ArXiv},
  year={2025},
  volume={abs/2508.10751},
  url={https://api.semanticscholar.org/CorpusID:280649795}
}

@misc{chen2024failuresselfconsistencymultistepreasoning,
      title={Two Failures of Self-Consistency in the Multi-Step Reasoning of LLMs}, 
      author={Angelica Chen and Jason Phang and Alicia Parrish and Vishakh Padmakumar and Chen Zhao and Samuel R. Bowman and Kyunghyun Cho},
      year={2024},
      eprint={2305.14279},
      archivePrefix={arXiv},
      primaryClass={cs.CL},
      url={https://arxiv.org/abs/2305.14279}, 
}

@misc{du2025optimizingtemperaturelanguagemodels,
      title={Optimizing Temperature for Language Models with Multi-Sample Inference}, 
      author={Weihua Du and Yiming Yang and Sean Welleck},
      year={2025},
      eprint={2502.05234},
      archivePrefix={arXiv},
      primaryClass={cs.LG},
      url={https://arxiv.org/abs/2502.05234}, 
}

@misc{qiang2024promptperturbationconsistencylearning,
      title={Prompt Perturbation Consistency Learning for Robust Language Models}, 
      author={Yao Qiang and Subhrangshu Nandi and Ninareh Mehrabi and Greg Ver Steeg and Anoop Kumar and Anna Rumshisky and Aram Galstyan},
      year={2024},
      eprint={2402.15833},
      archivePrefix={arXiv},
      primaryClass={cs.CL},
      url={https://arxiv.org/abs/2402.15833}, 
}

@inproceedings{Shi2024ATE,
  title={A Thorough Examination of Decoding Methods in the Era of LLMs},
  author={Chufan Shi and Haoran Yang and Deng Cai and Zhisong Zhang and Yifan Wang and Yujiu Yang and Wai Lam},
  booktitle={Conference on Empirical Methods in Natural Language Processing},
  year={2024},
  url={https://api.semanticscholar.org/CorpusID:267627384}
}

@article{ShurOfry2024GrowingAT,
  title={Growing a Tail: Increasing Output Diversity in Large Language Models},
  author={Michal Shur-Ofry and Bar Horowitz-Amsalem and Adir Rahamim and Yonatan Belinkov},
  journal={ArXiv},
  year={2024},
  volume={abs/2411.02989},
  url={https://api.semanticscholar.org/CorpusID:273821765}
}

@article{Holtzman2019TheCC,
  title={The Curious Case of Neural Text Degeneration},
  author={Ari Holtzman and Jan Buys and Li Du and Maxwell Forbes and Yejin Choi},
  journal={ArXiv},
  year={2019},
  volume={abs/1904.09751},
  url={https://api.semanticscholar.org/CorpusID:127986954}
}

@article{Renze2024TheEO,
  title={The Effect of Sampling Temperature on Problem Solving in Large Language Models},
  author={Matthew Renze and Erhan Guven},
  journal={ArXiv},
  year={2024},
  volume={abs/2402.05201},
  url={https://api.semanticscholar.org/CorpusID:267547769}
}

@article{Sutton1998ReinforcementLA,
  title={Reinforcement Learning: An Introduction},
  author={Richard S. Sutton and Andrew G. Barto},
  journal={IEEE Trans. Neural Networks},
  year={1998},
  volume={9},
  pages={1054-1054},
  url={https://api.semanticscholar.org/CorpusID:60035920}
}

@inproceedings{plappert2018parameter,
  title={Parameter Space Noise for Exploration},
  author={Plappert, Matthias and Houthooft, Rein and Dhariwal, Prafulla and Sidor, Szymon and Chen, Richard Y and Chen, Xi and Asfour, Tamim and Abbeel, Pieter and Andrychowicz, Marcin},
  booktitle={International Conference on Learning Representations},
  year={2018}
}

@article{shao2024deepseek,
   title={DeepSeek-R1 incentivizes reasoning in LLMs through reinforcement learning},
   volume={645},
   ISSN={1476-4687},
   url={http://dx.doi.org/10.1038/s41586-025-09422-z},
   DOI={10.1038/s41586-025-09422-z},
   number={8081},
   journal={Nature},
   publisher={Springer Science and Business Media LLC},
   author={Guo, Daya and Yang, Dejian and Zhang, Haowei and Song, Junxiao and Wang, Peiyi and Zhu, Qihao and Xu, Runxin and Zhang, Ruoyu and Ma, Shirong and Bi, Xiao and Zhang, Xiaokang and Yu, Xingkai and Wu, Yu and Wu, Z. F. and Gou, Zhibin and Shao, Zhihong and Li, Zhuoshu and Gao, Ziyi and Liu, Aixin and Xue, Bing and Wang, Bingxuan and Wu, Bochao and Feng, Bei and Lu, Chengda and Zhao, Chenggang and Deng, Chengqi and Ruan, Chong and Dai, Damai and Chen, Deli and Ji, Dongjie and Li, Erhang and Lin, Fangyun and Dai, Fucong and Luo, Fuli and Hao, Guangbo and Chen, Guanting and Li, Guowei and Zhang, H. and Xu, Hanwei and Ding, Honghui and Gao, Huazuo and Qu, Hui and Li, Hui and Guo, Jianzhong and Li, Jiashi and Chen, Jingchang and Yuan, Jingyang and Tu, Jinhao and Qiu, Junjie and Li, Junlong and Cai, J. L. and Ni, Jiaqi and Liang, Jian and Chen, Jin and Dong, Kai and Hu, Kai and You, Kaichao and Gao, Kaige and Guan, Kang and Huang, Kexin and Yu, Kuai and Wang, Lean and Zhang, Lecong and Zhao, Liang and Wang, Litong and Zhang, Liyue and Xu, Lei and Xia, Leyi and Zhang, Mingchuan and Zhang, Minghua and Tang, Minghui and Zhou, Mingxu and Li, Meng and Wang, Miaojun and Li, Mingming and Tian, Ning and Huang, Panpan and Zhang, Peng and Wang, Qiancheng and Chen, Qinyu and Du, Qiushi and Ge, Ruiqi and Zhang, Ruisong and Pan, Ruizhe and Wang, Runji and Chen, R. J. and Jin, R. L. and Chen, Ruyi and Lu, Shanghao and Zhou, Shangyan and Chen, Shanhuang and Ye, Shengfeng and Wang, Shiyu and Yu, Shuiping and Zhou, Shunfeng and Pan, Shuting and Li, S. S. and Zhou, Shuang and Wu, Shaoqing and Yun, Tao and Pei, Tian and Sun, Tianyu and Wang, T. and Zeng, Wangding and Liu, Wen and Liang, Wenfeng and Gao, Wenjun and Yu, Wenqin and Zhang, Wentao and Xiao, W. L. and An, Wei and Liu, Xiaodong and Wang, Xiaohan and Chen, Xiaokang and Nie, Xiaotao and Cheng, Xin and Liu, Xin and Xie, Xin and Liu, Xingchao and Yang, Xinyu and Li, Xinyuan and Su, Xuecheng and Lin, Xuheng and Li, X. Q. and Jin, Xiangyue and Shen, Xiaojin and Chen, Xiaosha and Sun, Xiaowen and Wang, Xiaoxiang and Song, Xinnan and Zhou, Xinyi and Wang, Xianzu and Shan, Xinxia and Li, Y. K. and Wang, Y. Q. and Wei, Y. X. and Zhang, Yang and Xu, Yanhong and Li, Yao and Zhao, Yao and Sun, Yaofeng and Wang, Yaohui and Yu, Yi and Zhang, Yichao and Shi, Yifan and Xiong, Yiliang and He, Ying and Piao, Yishi and Wang, Yisong and Tan, Yixuan and Ma, Yiyang and Liu, Yiyuan and Guo, Yongqiang and Ou, Yuan and Wang, Yuduan and Gong, Yue and Zou, Yuheng and He, Yujia and Xiong, Yunfan and Luo, Yuxiang and You, Yuxiang and Liu, Yuxuan and Zhou, Yuyang and Zhu, Y. X. and Huang, Yanping and Li, Yaohui and Zheng, Yi and Zhu, Yuchen and Ma, Yunxian and Tang, Ying and Zha, Yukun and Yan, Yuting and Ren, Z. Z. and Ren, Zehui and Sha, Zhangli and Fu, Zhe and Xu, Zhean and Xie, Zhenda and Zhang, Zhengyan and Hao, Zhewen and Ma, Zhicheng and Yan, Zhigang and Wu, Zhiyu and Gu, Zihui and Zhu, Zijia and Liu, Zijun and Li, Zilin and Xie, Ziwei and Song, Ziyang and Pan, Zizheng and Huang, Zhen and Xu, Zhipeng and Zhang, Zhongyu and Zhang, Zhen},
   year={2025},
   month=sep, pages={633–638} }

@inproceedings{Gupta2018MetaReinforcementLO,
  title={Meta-Reinforcement Learning of Structured Exploration Strategies},
  author={Abhishek Gupta and Russell Mendonca and Yuxuan Liu and P. Abbeel and Sergey Levine},
  booktitle={Neural Information Processing Systems},
  year={2018},
  url={https://api.semanticscholar.org/CorpusID:3418899}
}

@inproceedings{dang2025assessing,
  title     = {Assessing Diversity Collapse in Reasoning},
  author    = {Dang, Xingyu and Baek, Christina and Kolter, J. Zico and Raghunathan, Aditi},
  booktitle = {ICLR 2025 Workshop on Scaling Self-Improving Foundation Models without Human Supervision (SSI-FM)},
  year      = {2025},
  url       = {https://openreview.net/forum?id=AMiKsHLjQh}
}

@misc{amini2025betterestimationkullbackleiblerdivergence,
      title={Better Estimation of the Kullback--Leibler Divergence Between Language Models}, 
      author={Afra Amini and Tim Vieira and Ryan Cotterell},
      year={2025},
      eprint={2504.10637},
      archivePrefix={arXiv},
      primaryClass={cs.CL},
      url={https://arxiv.org/abs/2504.10637}, 
}

@article{ionides2008truncated,
  title   = {Truncated Importance Sampling},
  author  = {Ionides, Edward L.},
  journal = {Journal of Computational and Graphical Statistics},
  volume  = {17},
  number  = {2},
  pages   = {295--311},
  year    = {2008},
  doi     = {10.1198/106186008X320456}
}

@misc{gravell2021robustlearningbasedcontrolbootstrapped,
      title={Robust Learning-Based Control via Bootstrapped Multiplicative Noise}, 
      author={Benjamin Gravell and Tyler Summers},
      year={2021},
      eprint={2002.10069},
      archivePrefix={arXiv},
      primaryClass={cs.LG},
      url={https://arxiv.org/abs/2002.10069}, 
}

@article{Hollenstein2022ActionNI,
  title={Action Noise in Off-Policy Deep Reinforcement Learning: Impact on Exploration and Performance},
  author={Jakob J. Hollenstein and Sayantan Auddy and Matteo Saveriano and Erwan Renaudo and Justus H. Piater},
  journal={ArXiv},
  year={2022},
  volume={abs/2206.03787},
  url={https://api.semanticscholar.org/CorpusID:249461896}
}

@misc{huang2025qerlefficiencyquantizationenhanced,
      title={QeRL: Beyond Efficiency -- Quantization-enhanced Reinforcement Learning for LLMs}, 
      author={Wei Huang and Yi Ge and Shuai Yang and Yicheng Xiao and Huizi Mao and Yujun Lin and Hanrong Ye and Sifei Liu and Ka Chun Cheung and Hongxu Yin and Yao Lu and Xiaojuan Qi and Song Han and Yukang Chen},
      year={2025},
      eprint={2510.11696},
      archivePrefix={arXiv},
      primaryClass={cs.LG},
      url={https://arxiv.org/abs/2510.11696}, 
}

@article{osband2019deep,
  title   = {Deep Exploration via Randomized Value Functions},
  author  = {Osband, Ian and Van Roy, Benjamin and Russo, Daniel J. and Wen, Zheng},
  journal = {Journal of Machine Learning Research},
  volume  = {20},
  number  = {124},
  pages   = {1--62},
  year    = {2019}
}

@inproceedings{russo2019worstcase,
  title     = {Worst-Case Regret Bounds for Exploration via Randomized Value Functions},
  author    = {Russo, Daniel},
  booktitle = {Advances in Neural Information Processing Systems (NeurIPS)},
  volume    = {32},
  pages     = {14410--14420},
  year      = {2019}
}

@inproceedings{aravindan2021stateaware,
  title     = {State-Aware Variational Thompson Sampling for Deep Q-Networks},
  author    = {Aravindan, Siddharth and Lee, Wee Sun},
  booktitle = {AAMAS '21: 20th International Conference on Autonomous Agents and Multiagent Systems},
  pages     = {124--132},
  publisher = {ACM},
  year      = {2021},
  doi       = {10.5555/3463952.3463973}
}

@inproceedings{he2024olympiadbench,
  title={Olympiadbench: A challenging benchmark for promoting agi with olympiad-level bilingual multimodal scientific problems},
  author={He, Chaoqun and Luo, Renjie and Bai, Yuzhuo and Hu, Shengding and Thai, Zhen and Shen, Junhao and Hu, Jinyi and Han, Xu and Huang, Yujie and Zhang, Yuxiang and others},
  booktitle={Proceedings of the 62nd Annual Meeting of the Association for Computational Linguistics (Volume 1: Long Papers)},
  pages={3828--3850},
  year={2024}
}

@article{lewkowycz2022solving,
  title={Solving quantitative reasoning problems with language models},
  author={Lewkowycz, Aitor and Andreassen, Anders and Dohan, David and Dyer, Ethan and Michalewski, Henryk and Ramasesh, Vinay and Slone, Ambrose and Anil, Cem and Schlag, Imanol and Gutman-Solo, Theo and others},
  journal={Advances in neural information processing systems},
  volume={35},
  pages={3843--3857},
  year={2022}
}
\bibliographystyle{icml2026}

\newpage
\appendix
\onecolumn
\section{Related works for Reinforcement Learning for Reasoning with Verifiable Rewards}
\label{sec:app_related_work}
The integration of Reinforcement Learning (RL) into the post-training pipeline of Large Language Models (LLMs) has become a standard paradigm for enhancing performance in domains with objective ground-truth, such as mathematics and coding \cite{ouyang2022training, wang2024mathvista}. Early approaches primarily utilized Proximal Policy Optimization (PPO) \cite{schulman2017proximal} to align models with reward signals derived from unit tests or symbolic solvers. More recently, Group Relative Policy Optimization (GRPO) \cite{shao2024deepseek} has gained prominence by eliminating the need for a separate value network, instead normalizing advantages within a group of sampled outputs. This efficiency has enabled massive-scale RL training, exemplified by models like DeepSeek-R1, which demonstrate emergent reasoning capabilities solely through reinforcement signals. However, emerging evidence suggests that current RLVR pipelines may face an \emph{exploration ceiling}. Recent analyses~\cite{yue2025does, wu2026invisibleleashrlvrescape} show that RLVR predominantly improves \emph{selection efficiency} among pre-existing trajectories, with limited ability to generate genuinely new reasoning modes, leaving the learned policy largely constrained by the base model’s pretraining distribution and exposing a critical exploration bottleneck.
\section{Training  experiment setting}
\label{app:train_detail}
\begin{table}[h]
\centering
\caption{Hyperparameters used for SimpleRL-Zoo models.}
\label{tab:simplerl_params}
\begin{tabular}{l|c}
\toprule
\textbf{Settings} & \textbf{SimpleRL-Zoo} \\
\midrule
Framework & verl~\cite{sheng2024hybridflow} \\
Prompt Batch Size & 512 \\
Mini-batch Size & 256 \\
\# Policy Rollout $G$ & 8 \\
Max Rollout Length & 8,192 \\
Max Generation Length (Eval) & 16,384 \\
Clip Ratio & 0.2 \\
KL Loss Coefficient & $1 \times 10^{-4}$ \\
Training Temperature & 1.0 \\
Evaluation Temperature & 0.9 \\
Adaptive Noise Update Step $\beta$ & 1.01 \\
Adaptive Noise Scale Range & $[0.8\sigma_{\text{init}}, 1.2\sigma_{\text{init}}]$ \\
Target Divergence $\delta$ & 0.03 \\
$C$ in TIS Equation~\ref{eq:tis_c}&10\\
\bottomrule
\end{tabular}
\end{table}
\section{Compute overhead.}
\label{sec:computation_overhead}
Variant~II uses two probe generations per query under \(\pi_{\theta}\) to estimate semantic similarity and self-certainty before sampling the \(G\) training rollouts under \(\pi_{\tilde{\theta}}\).
With \(G=8\), a naive upper-bound estimate suggests \(2/8=0.25\) additional generations time; if on-policy rollout generation occupies at least \(70\%\) of wall-clock training time, this would imply an expected slowdown of roughly \(0.25\times 0.70 \approx 0.17\) (i.e., \(\approx 17\%\)).
Empirically, however, we observe a smaller end-to-end throughput reduction of per-iteration \(\approx 8\%\) relative to fixed-\(\sigma\) PSN under identical hardware, batch size, and maximum generation length.
We attribute the gap to \emph{generation-time imbalance}~\cite{he2025historyrhymesacceleratingllm}: in batched/parallel decoding, shorter sequences finish earlier and GPUs can become partially idle while waiting for the longest sequences (stragglers) to complete.
Adding a small number of short probe generations reduces the variance in per-step generation workload and mitigates tail-latency effects, leading to a wall-clock overhead that is lower than the token-count-based estimate.
\clearpage
\onecolumn  
\section{Result of Qwen3-4B-Base}
\begin{table*}[h]
\centering
\setlength{\tabcolsep}{6pt}
\renewcommand{\arraystretch}{1.08}
\resizebox{\textwidth}{!}{
\begin{tabular}{l ccc ccc ccc}
\toprule
\multirow{2}{*}{\textbf{Benchmark}} &
\multicolumn{3}{c}{\textbf{Clean ($\sigma=0$)}} &
\multicolumn{3}{c}{\textbf{PSN ($\sigma=0.001$)}} &
\multicolumn{3}{c}{\textbf{Gap (PSN$-$Clean)}} \\
\cmidrule(lr){2-4}\cmidrule(lr){5-7}\cmidrule(lr){8-10}
& \textbf{pass@1} & \textbf{pass@2} & \textbf{pass@256}
& \textbf{pass@1} & \textbf{pass@2} & \textbf{pass@256}
& \textbf{pass@1} & \textbf{pass@2} & \textbf{pass@256} \\
\midrule
AIME 24        & 19.4\% & 25.7\% & 62.8\% & 17.4\% & 23.6\% & 65.5\% & -2.0 & -2.1 & +2.7 \\
AIME 25        & 15.5\% & 20.6\% & 58.9\% & 14.7\% & 20.0\% & 62.6\% & -0.8 & -0.6 & +3.7 \\
Minerva-Math   & 32.2\% & 39.3\% & 63.6\% & 32.0\% & 40.7\% & 64.8\% & -0.2 & +1.4 & +1.2 \\
OlympiadBench  & 44.4\% & 52.9\% & 77.1\% & 40.6\% & 50.8\% & 77.4\% & -3.8 & -2.1 & +0.3 \\
\bottomrule
\end{tabular}
}
\caption{\textbf{Qwen3-4B-Base detailed results (clean vs.\ PSN).} Per-benchmark pass@$k$ (\%) for Qwen3-4B-Base comparing clean decoding ($\sigma=0$) and PSN ($\sigma=0.001$; MLP-only perturbation). PSN exhibits the typical exploration--exploitation trade-off: slightly lower low-budget reliability (pass@1/2) but an expanded high-budget capability boundary (pass@256), especially on harder AIME benchmarks.}
\label{tab:q2_qwen3_4bbase_detail}
\end{table*}
\clearpage

\section{Detailed performance of training time temperature scaling}
\begin{figure}[h!]
    \centering
    \includegraphics[width=0.95\linewidth]{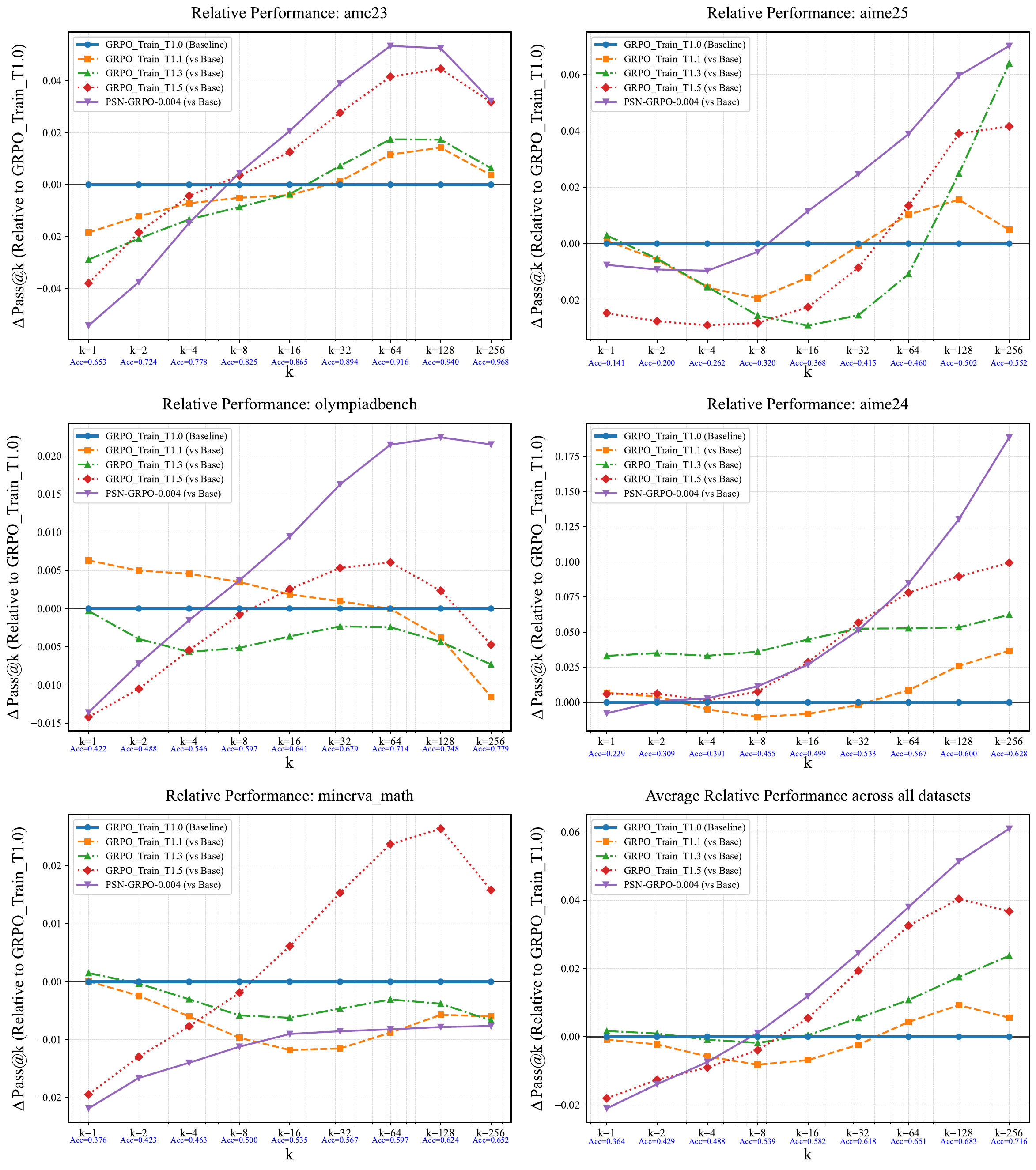}
    \caption{Performance across benchmarks. We do not plot T=1.7 in these plots since its overall performance collapsed as shown in Figure~\ref{fig:training_temp_scaling}.}
    \label{fig:training_temp_scaling_detail}
\end{figure}
\clearpage
\section{More experiment result of evaluation temperature scaling}
\begin{figure}[h!]
    \centering
    \includegraphics[width=0.9\linewidth]{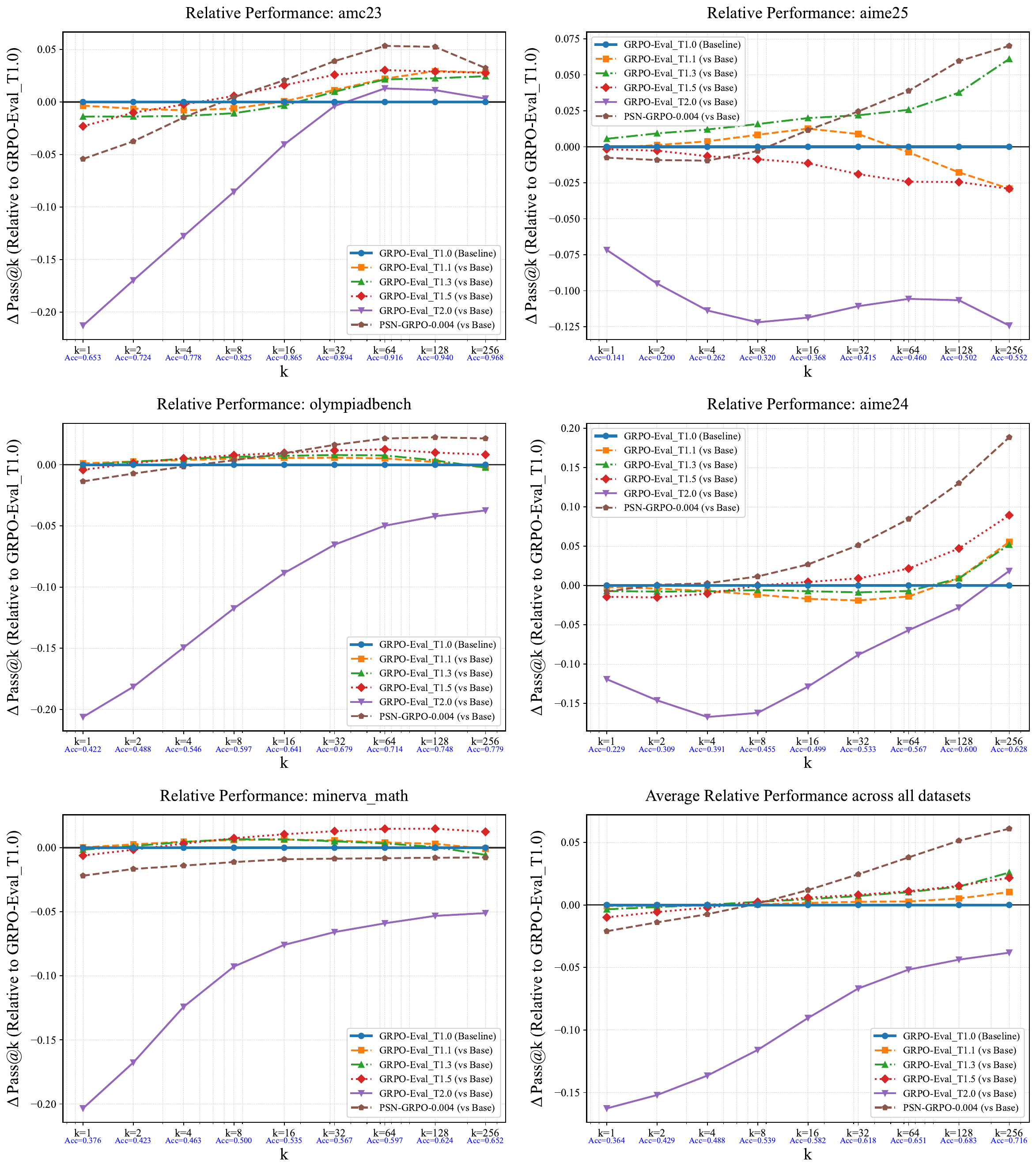}
\caption{Detailed performance analysis across five datasets. }
    \label{fig:evaluation_noise_scaling_detail}
\end{figure}
\clearpage

\section{More detailed experiment result of evaluation temperature scaling}
\begin{figure}[h!]
    \centering
    \includegraphics[width=0.95\linewidth]{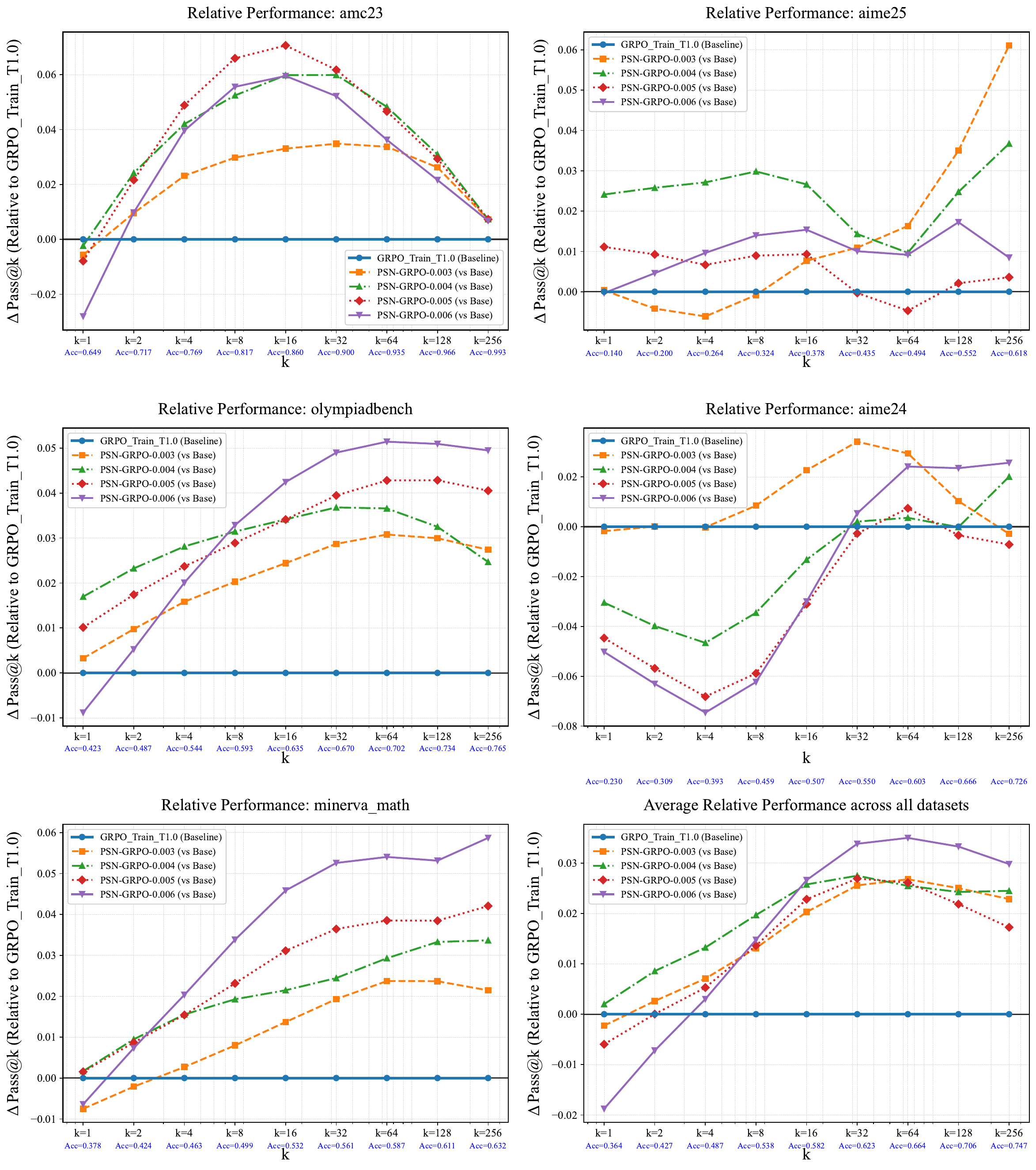}
\caption{Detailed performance analysis across five datasets. In general, a larger noise scale $\sigma$ tends to yield higher pass@$k$ for large $k$. Specifically, on datasets of moderate difficulty (e.g., OlympiadBench and Minerva Math), increased $\sigma$ improves reasoning capability boundary, yield higher pass@$k$ for large $k$. Conversely, on more challenging benchmarks such as AIME24 and AIME25, a larger noise scale tends to degrade overall performance.}
    \label{fig:more_q5}
\end{figure}
\clearpage

\section{Detailed performance of where to inject noise}
\begin{figure}[h!]
    \centering
    \includegraphics[width=0.95\linewidth]{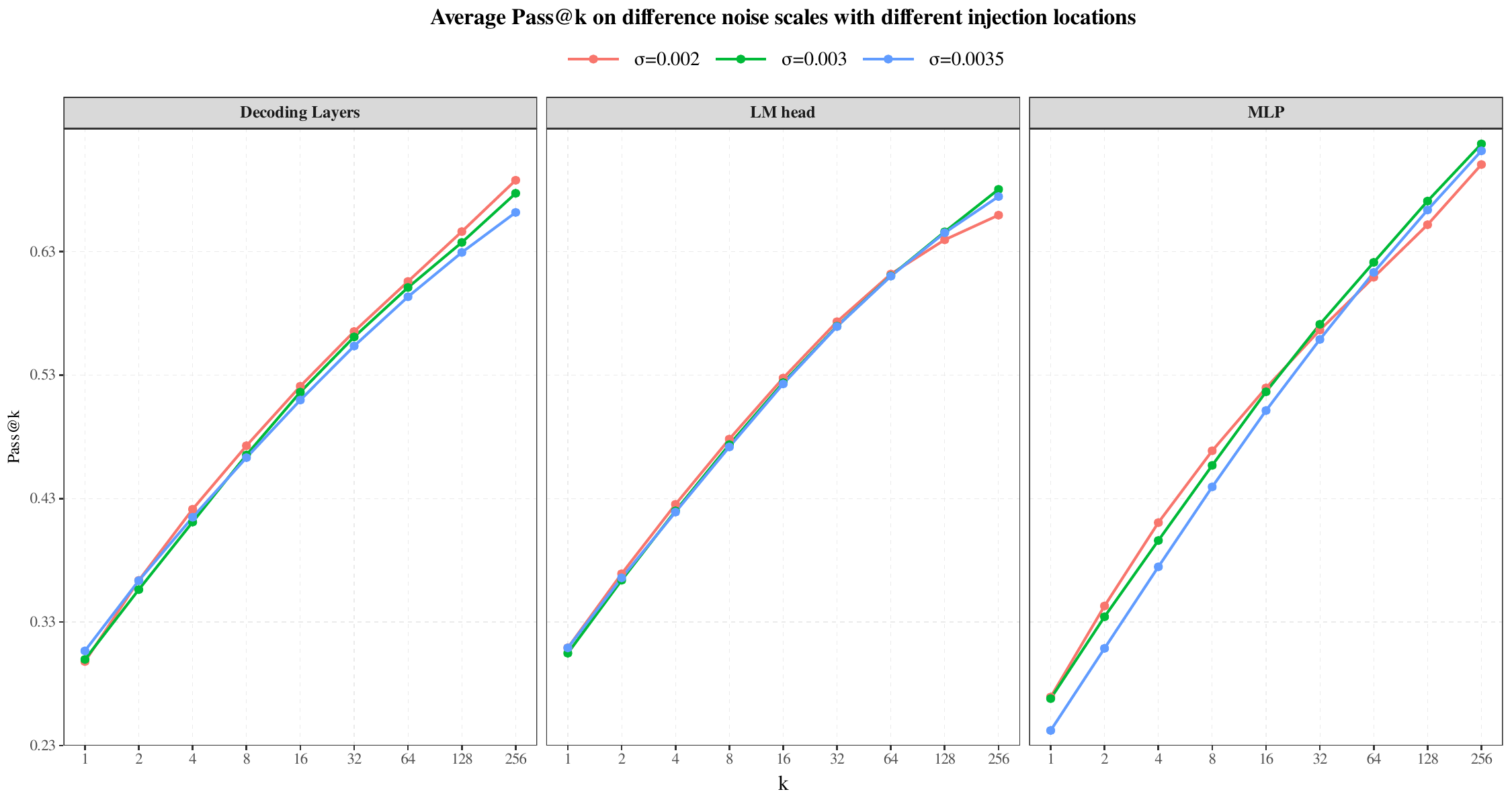}
    \caption{Average Pass@k under parameter-space noise injected whole  transformer layers, the \texttt{lm\_head}, or all MLP sublayers, sweeping noise scale $\sigma$. MLP injection yields the strongest and most consistent improvements, especially at large $k$.}
    \label{fig:where_inject_noise_detail}
\end{figure}
\clearpage

\section{Detailed performance of adaptive noise}
\begin{table}[h]
\centering
\setlength{\tabcolsep}{5pt} 
\resizebox{0.8\linewidth}{!}{
\begin{tabular}{l l cc cc cc}
\toprule
\multirow{2}{*}{\textbf{Dataset}} & \multirow{2}{*}{\textbf{Method}} & \multicolumn{2}{c}{\textbf{Pass@2}} & \multicolumn{2}{c}{\textbf{Pass@4}} & \multicolumn{2}{c}{\textbf{Pass@256}} \\
\cmidrule(lr){3-4} \cmidrule(lr){5-6} \cmidrule(lr){7-8}
& & \textbf{Score} & \textbf{Gap} & \textbf{Score} & \textbf{Gap} & \textbf{Score} & \textbf{Gap} \\
\midrule
\multirow{4}{*}{AMC 23} &  PSN Var-II & \underline{\textbf{74.3\%}} & - & \underline{\textbf{81.8\%}} & - & 99.6\% & - \\
& GRPO Train & 71.7\% & \textcolor{darkgreen}{+2.6\%} & 76.9\% & \textcolor{darkgreen}{+4.8\%} & 99.3\% & \textcolor{darkgreen}{+0.4\%} \\
& PSN Fixed & 74.1\% & \textcolor{darkgreen}{+0.2\%} & 81.1\% & \textcolor{darkgreen}{+0.6\%} & \underline{\textbf{100.0\%}} & \textcolor{orange}{-0.4\%} \\
& PSN Var-I & 72.6\% & \textcolor{darkgreen}{+1.7\%} & 79.2\% & \textcolor{darkgreen}{+2.6\%} & 100.0\% & \textcolor{orange}{-0.4\%} \\
\midrule
\multirow{4}{*}{Olympiad} &  PSN Var-II & \underline{\textbf{51.3\%}} & - & \underline{\textbf{57.7\%}} & - & \underline{\textbf{81.1\%}} & - \\
& GRPO Train & 48.7\% & \textcolor{darkgreen}{+2.6\%} & 54.4\% & \textcolor{darkgreen}{+3.3\%} & 76.5\% & \textcolor{darkgreen}{+4.7\%} \\
& PSN Fixed & 51.1\% & \textcolor{darkgreen}{+0.3\%} & 57.2\% & \textcolor{darkgreen}{+0.5\%} & 78.9\% & \textcolor{darkgreen}{+2.2\%} \\
& PSN Var-I & 50.3\% & \textcolor{darkgreen}{+1.0\%} & 56.4\% & \textcolor{darkgreen}{+1.3\%} & 78.6\% & \textcolor{darkgreen}{+2.6\%} \\
\midrule
\multirow{4}{*}{AIME 25} &  PSN Var-II & \underline{\textbf{22.9\%}} & - & \underline{\textbf{29.2\%}} & - & \underline{\textbf{65.8\%}} & - \\
& GRPO Train & 20.0\% & \textcolor{darkgreen}{+2.9\%} & 26.4\% & \textcolor{darkgreen}{+2.8\%} & 61.8\% & \textcolor{darkgreen}{+4.0\%} \\
& PSN Fixed & 22.6\% & \textcolor{darkgreen}{+0.3\%} & 29.1\% & \textcolor{darkgreen}{+0.1\%} & 65.4\% & \textcolor{darkgreen}{+0.3\%} \\
& PSN Var-I & 20.7\% & \textcolor{darkgreen}{+2.2\%} & 26.9\% & \textcolor{darkgreen}{+2.3\%} & 55.1\% & \textcolor{darkgreen}{+10.7\%} \\
\midrule
\multirow{4}{*}{AIME 24} &  PSN Var-II & 28.1\% & - & 37.1\% & - & \underline{\textbf{81.7\%}} & - \\
& GRPO Train & \underline{\textbf{30.9\%}} & \textcolor{orange}{-2.8\%} & \underline{\textbf{39.3\%}} & \textcolor{orange}{-2.2\%} & 72.6\% & \textcolor{darkgreen}{+9.1\%} \\
& PSN Fixed & 27.0\% & \textcolor{darkgreen}{+1.1\%} & 34.6\% & \textcolor{darkgreen}{+2.4\%} & 74.6\% & \textcolor{darkgreen}{+7.1\%} \\
& PSN Var-I & 25.5\% & \textcolor{darkgreen}{+2.6\%} & 32.9\% & \textcolor{darkgreen}{+4.2\%} & 75.0\% & \textcolor{darkgreen}{+6.7\%} \\
\midrule
\multirow{4}{*}{Minerva} &  PSN Var-II & \underline{\textbf{43.7\%}} & - & 47.4\% & - & \underline{\textbf{69.1\%}} & - \\
& GRPO Train & 42.4\% & \textcolor{darkgreen}{+1.4\%} & 46.3\% & \textcolor{darkgreen}{+1.0\%} & 63.2\% & \textcolor{darkgreen}{+5.9\%} \\
& PSN Fixed & 43.3\% & \textcolor{darkgreen}{+0.4\%} & 47.9\% & \textcolor{orange}{-0.5\%} & 66.6\% & \textcolor{darkgreen}{+2.5\%} \\
& PSN Var-I & 43.5\% & \textcolor{darkgreen}{+0.3\%} & \underline{\textbf{48.0\%}} & \textcolor{orange}{-0.7\%} & 66.7\% & \textcolor{darkgreen}{+2.4\%} \\
\midrule
\multirow{4}{*}{Average} & PSN Var-II & \underline{\textbf{44.1\%}} & - & \underline{\textbf{50.6\%}} & - & \underline{\textbf{79.5\%}} & - \\
& GRPO Train & 42.7\% & \textcolor{darkgreen}{+1.3\%} & 48.7\% & \textcolor{darkgreen}{+1.9\%} & 74.7\% & \textcolor{darkgreen4}{+4.8\%} \\
& PSN Fixed & 43.6\% & \textcolor{darkgreen}{+0.5\%} & 50.0\% & \textcolor{darkgreen}{+0.6\%} & 77.1\% & \textcolor{darkgreen}{+2.4\%} \\
& PSN Var-I & 42.5\% & \textcolor{darkgreen}{+1.5\%} & 48.7\% & \textcolor{darkgreen2}{+1.9\%} & 75.1\% & \textcolor{darkgreen4}{+4.4\%} \\
\bottomrule
\end{tabular}
}
\caption{\textbf{Performance comparison across mathematical benchmarks.} We report Pass@$k$ scores (\%) for our adaptive variants (PSN Var-I/II) against the standard GRPO baseline and Fixed-Noise PSN ($\sigma=0.004$). \textbf{Gap} denotes the absolute percentage point improvement of \textit{PSN Var-II} over the compared method. Notably, the non-real-time schedule (Var-I) hurts performance due to feedback latency. }
\label{tab:adaptive_noise_advantage_detail}
\end{table}
\clearpage
\definecolor{case-orange-bg}{RGB}{255, 246, 230}
\definecolor{case-orange-border}{RGB}{234, 158, 87}
\definecolor{case-pink-bg}{RGB}{255, 235, 245}
\definecolor{case-pink-border}{RGB}{210, 100, 150}
\definecolor{case-blue-bg}{RGB}{235, 240, 255}
\definecolor{case-blue-border}{RGB}{100, 120, 240}
\definecolor{darkgreen}{RGB}{0, 100, 0}

\newtcolorbox{casequestion}[1]{
    enhanced, colback=case-orange-bg, colframe=case-orange-border, 
    fonttitle=\bfseries, coltitle=black, title={#1},
    attach boxed title to top left={yshift=-2mm, xshift=3mm},
    boxed title style={colback=case-orange-border, sharp corners, frame hidden},
    top=4mm, before skip=10pt, after skip=0pt
}

\newtcolorbox{baselinebox}[1]{
    enhanced, colback=case-orange-bg, colframe=case-orange-border, 
    fonttitle=\bfseries, coltitle=black, title={#1},
    attach boxed title to top left={yshift=-2mm, xshift=3mm},
    boxed title style={colback=case-orange-border, sharp corners, frame hidden},
    top=4mm, before skip=12pt, after skip=2pt
}

\newtcolorbox{psnbox}[1]{
    enhanced, colback=case-pink-bg, colframe=case-pink-border, 
    fonttitle=\bfseries, coltitle=black, title={#1},
    attach boxed title to top left={yshift=-2mm, xshift=3mm},
    boxed title style={colback=case-pink-border, sharp corners, frame hidden},
    top=4mm, before skip=12pt
}

\newtcolorbox{caseverify}{
    colback=case-blue-bg, colframe=case-blue-border, arc=3pt, boxrule=0.8pt,
    left=8pt, right=8pt, top=4pt, bottom=4pt, fontupper=\small,
    before skip=4pt, after skip=8pt
}


\section{Detailed example of PSN-GRPO discovers qualitatively new solution strategies}
\label{sec:detailed_new_solution}

\begin{casequestion}{Question (AIME I 2024, Problem 6)}
    Each vertex of a regular octagon is independently colored either red or blue with equal probability. The probability that the octagon can then be rotated so that all of the blue vertices end up at positions where there were originally red vertices is $\frac{m}{n}$, where $m$ and $n$ are relatively prime positive integers. What is $m+n$?
\end{casequestion}

\begin{baselinebox}{\textcolor{red}{\ding{55}} Baseline Error Logic 1: Invariance Trap}
    The model incorrectly simplifies the problem by assuming the coloring must be invariant under a non-trivial rotation. 
    \begin{itemize}[leftmargin=1.5em, nosep]
        \item It identifies $k \in \{0, 4, 8\}$ as the only counts allowing rotation.
        \item For $k=4$, it only counts 2 symmetric patterns (alternating positions $\{1,3,5,7\}$ and $\{2,4,6,8\}$).
    \end{itemize}
    Total colorings = $1 (k=0) + 2 (k=4) + 1 (k=8) = 4$.
    \begin{caseverify}
        \textbf{Diversity Collapse:} The policy has collapsed to a small subset of globally symmetric solutions. It fails to recognize that ``rotatable to red'' simply requires the set of blue vertices $B$ to satisfy $B \cap R_\theta(B) = \emptyset$ for at least one $\theta$.
    \end{caseverify}
\end{baselinebox}

\begin{baselinebox}{\textcolor{red}{\ding{55}} Baseline Error Logic 2: Local Symmetry Constraint}
    For $k=2$, the model argues that if vertices are at positions $\{1, 3\}$, they cannot be rotated because ``the $90^\circ$ separation does not align with the octagon's fundamental $45^\circ$ or $180^\circ$ symmetry axis.'' It erroneously prunes all non-opposite pairs.
    \begin{caseverify}
        \textbf{Exploration Ceiling:} This demonstrates a failure to simulate even a single-step $45^\circ$ rotation ($1 \rightarrow 2, 3 \rightarrow 4$). The model incorrectly filters out valid configurations that do not satisfy its rigid internal symmetry priors, leading to a massive undercount.
    \end{caseverify}
\end{baselinebox}

\begin{baselinebox}{\textcolor{red}{\ding{55}} Baseline Error Logic 3: Counting Pruning Error}
    For $k=3$ and $k=4$, the model bypasses enumeration entirely. It claims that since $k \ge 3$ creates ``too many constraints on adjacent slots,'' it is statistically impossible for a rotation to map all blue vertices to red positions unless the distribution is perfectly uniform (i.e., the $k=4$ case in Error 1).
    \begin{caseverify}
        \textbf{Reasoning Failure:} The model uses pigeonhole-style heuristics incorrectly. It fails to explore asymmetric but valid sparse placements (e.g., $k=3$ with no two adjacent). PSN-GRPO finds 40 and 46 cases for $k=3, 4$, which the baseline misses due to premature search pruning.
    \end{caseverify}
\end{baselinebox}

\begin{psnbox}{\textcolor{darkgreen}{\ding{52}} PSN-GRPO Answer (Ours)}
    PSN-GRPO identifies the problem's core requirement: finding all $B \subseteq V$ such that $\exists \theta \in \{45^\circ, \dots, 315^\circ\}, R_\theta(B) \cap B = \emptyset$.

    \textbf{Comprehensive Strategy Discovery:}
    \begin{itemize}[leftmargin=1.5em, nosep]
        \item \textbf{$k=0, 1$}: All $\binom{8}{0} + \binom{8}{1} = 9$ colorings work.
        \item \textbf{$k=2$}: Valid if not adjacent ($B=\{1,2\}$ fails for all $\theta$). Valid = $\binom{8}{2} - 8 = 20$.
        \item \textbf{$k=3$}: PSN explores non-symmetric triplets. It correctly excludes triplets like $\{1,2,3\}$ and $\{1,2,x\}$ where all rotations are blocked. Valid = $\binom{8}{3} - 16 = 40$.
        \item \textbf{$k=4$}: Discovers 46 valid cases by maintaining diverse reasoning paths that avoid the ``greedy trap'' of symmetry.
    \end{itemize}
    
    \begin{caseverify}
        \textbf{Strategic Discovery:} PSN-GRPO's parameter-space perturbation allows it to sustain multiple valid reasoning trajectories for complex counts ($k=3, 4$), eventually summing to 115 favorable colorings.
    \end{caseverify}
    
    Total Probability = $115/256$. Result: $m+n = 115 + 256 = \mathbf{371}$.
\end{psnbox}

\end{document}